# MEASURING AXIOMATIC SOUNDNESS OF COUNTERFACTUAL IMAGE MODELS

**Miguel Monteiro**[1,†] **Fabio De Sousa Ribeiro**[1,†] **Nick Pawlowski**[2] **Daniel C. Castro**[1,2] **Ben Glocker**[1]
[1]Imperial College London, [2]Microsoft Research Cambridge. [†]Joint first authors
`{miguel.monteiro,f.de-sousa-ribeiro,b.glocker}@imperial.ac.uk`

## ABSTRACT

We present a general framework for evaluating image counterfactuals. The power and flexibility of deep generative models make them valuable tools for learning mechanisms in structural causal models. However, their flexibility makes counterfactual identifiability impossible in the general case. Motivated by these issues, we revisit Pearl's axiomatic definition of counterfactuals to determine the necessary constraints of any counterfactual inference model: *composition*, *reversibility*, and *effectiveness*. We frame counterfactuals as functions of an input variable, its parents, and counterfactual parents and use the axiomatic constraints to restrict the set of functions that could represent the counterfactual, thus deriving distance metrics between the approximate and ideal functions. We demonstrate how these metrics can be used to compare and choose between different approximate counterfactual inference models and to provide insight into a model's shortcomings and trade-offs.

## 1 INTRODUCTION

Faithfully answering counterfactual queries is a key challenge in representation learning and a cornerstone for aligning machine intelligence and human reasoning. While significant advances have been made in causal representation learning, enabling approximate counterfactual inference, there is surprisingly little methodology available to assess, measure, and quantify the quality of these models.

The structural causal model (SCM) is a mathematical tool that describes causal systems. It offers a convenient computational framework for operationalising causal and counterfactual inference (Pearl, 2009). An SCM is a set of functional assignments (called *mechanisms*) that represent the relationship between a variable, its direct causes (called *parents*) and all other unaccounted sources of variation (called *exogenous noise*). In SCMs, we assume that the mechanisms are algorithmically independent of each other. Further, in Markovian SCMs, which we can represent by DAGs, we assume that the exogenous noise variables are statistically independent of each other (Peters et al., 2017). From here on out, by SCM, we mean Markovian SCM. When the functional form of a mechanism is unknown, learning it from data is a prerequisite for answering counterfactual queries (Bareinboim et al., 2022).

In the context of high-dimensional observations, such as images, the power and flexibility of deep generative models make them indispensable tools for learning the mechanisms of an SCM. However, these same benefits make model identifiability impossible in the general case (Locatello et al., 2020a; Khemakhem et al., 2020), which can cause entanglement of causal effects (Pawlowski, 2021) and lead to poor approximations of the causal quantities of interest. Regardless, even if the model or counterfactual query is unidentifiable, we can still measure the quality of the counterfactual approximation (Pearl, 2010). However, evaluating image-counterfactual models is challenging without access to observed counterfactuals, which is unrealistic in most real-world scenarios.

In this paper, we focus on what constraints a counterfactual inference model must satisfy and how we can use them to measure the model's *soundness* without having access to observed counterfactuals or the SCM that generated the data. We begin by framing mechanisms as functional assignments that directly translate an observation into a counterfactual, given its parents and counterfactual parents. Next, we use Galles & Pearl (1998)'s axiomatic definition of counterfactual to restrict the space of possible functions that can represent a mechanism. From these constraints, we derive a set of metrics which we can use to measure the soundness of any arbitrary black-box counterfactual inference engine. Lastly, we show how simulated interventions can mitigate estimation issues due to confounding.

## 2 RELATED WORK

Representation learning aims to capture semantically meaningful disentangled factors of variation in the data. Arguably, these representations can provide interpretability, reduced sample complexity, and improved generalisation (Bengio et al., 2013). From a causal perspective, these factors should represent the parents of a variable in the SCM (Schölkopf et al., 2021). Although there has been extensive research into unsupervised disentanglement (Higgins et al., 2017; Burgess et al., 2018; Kim & Mnih, 2018; Chen et al., 2018; Kumar et al., 2018; Peebles et al., 2020), recent results Locatello et al. (2020a) reaffirm the impossibility of this task since the true causal generative model is not identifiable by observing a variable in isolation (Peters et al., 2017). In contrast, supervised disentanglement, where we observe the variable's parents, and weakly supervised disentanglement, where we observe "real" counterfactuals, can lead to causally identifiable generative models (Locatello et al., 2020a).

The integration of causal considerations has led to the emerging field of causal representation learning (Schölkopf et al., 2021). In the supervised setting, extensive research has been conducted in adapting deep models for individualised treatment effect estimation (Louizos et al., 2017; Yoon et al., 2018; Shi et al., 2019; Madras et al., 2019; Jesson et al., 2020; Yang et al., 2021). Notably, Louizos et al. (2017) use deep latent variable models to estimate individualised and population-level treatment effects. Yang et al. (2021) use deep latent variable models for learning to transform independent exogenous factors into endogenous causes that correspond to causally related concepts in the data. In the weakly-supervised setting, recent work has focused on using observations of "real"-counterfactuals instead of a variable's parents to obtain disentangled representations (Hosoya, 2019; Bouchacourt et al., 2018; Shu et al., 2020; Locatello et al., 2020b). Besserve et al. (2020; 2021) show how relaxing identifiability constraints can lead to some degree of identifiability in unsupervised settings.

In the context of image counterfactuals, Pawlowski et al. (2020) demonstrate how to jointly model all the functional assignments in an SCM using deep generative models. Despite presenting a general theory for any generative model, the authors implement only VAEs (Kingma & Welling, 2014; Rezende et al., 2014) and normalising flows (Papamakarios et al., 2021), which Dash et al. (2022) complement by using GANs (Goodfellow et al., 2014). Sanchez & Tsaftaris (2022) use diffusion models for the counterfactual estimation. Van Looveren & Klaise (2021) use class prototypes for finding interpretable counterfactual explanations. Sauer & Geiger (2021) use a deep network to disentangle object shape, object texture and background in natural images. Parascandolo et al. (2018) retrieve a set of independent mechanisms from a set of transformed data points in an unsupervised manner using multiple competing models. Additionally, many image-to-image translation models can be considered informal counterfactual inference engines (Isola et al., 2017; Zhu et al., 2017; Liu et al., 2017; Choi et al., 2018; Hoffman et al., 2018; Li et al., 2021).

The flexibility of deep models makes them susceptible to learning shortcuts (Geirhos et al., 2020). Consequently, when the data is biased, the effects of the parents can become entangled (Pawlowski, 2021; Rissanen & Marttinen, 2021). These issues create identifiability problems even when accounting for causality in representation learning. Simulating interventions through data augmentation or resampling can be used to debias the data (Ilse et al., 2021; Kügelgen et al., 2021; An et al., 2021). In a closely related field, research has focused on learning from biased data (Nam et al., 2020) or how to become invariant to a protected/spurious attribute (Kim et al., 2019). Janzing & Schölkopf (2010) show how an algorithmic perspective allows causal inference with only one observation.

## 3 METHODS

Generating counterfactuals is commonly performed in multiple steps. First, we abduct the exogenous noise from the observation and its parents. Second, we act on some parents. Finally, we use a generative model to map the exogenous noise and the counterfactual parents back to the observation space. For deep models, true abduction is impossible in the general case, and identifiability issues are ubiquitous (Peters et al., 2017; Locatello et al., 2020a; Khemakhem et al., 2020). There can exist multiple models capable of generating the data, and the true causal model cannot be identified from data alone. We argue that viewing counterfactual inference engines as black-boxes, where we focus on what properties the model's output must obey rather than the model's theoretical shortcomings, leads us to a set of actionable and principled model constraints. While a full causal model of the data generation process is necessary to create new samples from a joint distribution, in many applications, we are only interested in generating counterfactuals for real observations. In this case, we can directly model the mapping between observation and counterfactual. We call this mapping a counterfactual function in which we consider the abduction to be an implicit process rather than an explicit step in the counterfactual inference.

### 3.1 COUNTERFACTUAL FUNCTIONS

Let $x$ be a random variable (i.e., the observation) with parents $\mathbf{pa}$, $x^*$ a counterfactual of $x$ with parents $\mathbf{pa}^*$, and $\epsilon$ the exogenous noise variable pertaining to $x$. The functional assignment for $x$ is given by $x \coloneqq \mathrm{g}(\epsilon, \mathbf{pa})$, and for its counterfactual by $x^* \coloneqq \mathrm{g}(\epsilon, \mathbf{pa}^*)$. Conceptually, counterfactual reasoning is a three-step process: 1) Abduction: infer the exogenous noise from the observation and its parents; 2) Action: intervene on the targeted parents; 3) Prediction: propagate the effect of the intervention through the modified model to generate the counterfactual (Pearl, 2009). However, when the values for all the counterfactual parents are known, abduction, action and prediction do not need to be performed sequentially but can be formulated as a single functional assignment taking as arguments the observation, the parents and the counterfactual parents, as formalised in the following.

The abduction process involves inverting the mechanism with respect to the exogenous noise $\epsilon \coloneqq \mathrm{g}^{-1}(x, \mathbf{pa})$. In general, the mechanism is not invertible since there might be several possible counterfactuals corresponding to the same observation. In other words, the inverse mapping is not deterministic since an observation induces a distribution over the exogenous noise, which induces a distribution over possible counterfactuals: $\epsilon \sim P(\epsilon|x, \mathbf{pa}) \implies x^* \sim P(x^*|x, \mathbf{pa})$. Equivalently, we can formulate abduction as a deterministic functional assignment $\epsilon \coloneqq \mathrm{abduct}(x, \mathbf{pa})$ where the abduction function is drawn from a distribution over functions $\mathrm{abduct} \sim P(\mathrm{abduct})$. We can rewrite the functional assignment for the counterfactual as $x^* \coloneqq \mathrm{g}(\mathrm{abduct}(x, \mathbf{pa}), \mathbf{pa}^*)$ and obtain an equivalent expression by introducing a new function with the same arguments $x^* \coloneqq \mathrm{f}(x, \mathbf{pa}, \mathbf{pa}^*)$, where $\mathrm{f} \sim P(\mathrm{f})$. We call these functions counterfactual functions and denote the abduction as implicit since a value for the exogenous noise $\epsilon$ is never explicitly produced. Notice how abduct and f do not depend on the data due to the independence of cause and mechanism (Peters et al., 2017, Sec. 2.1).

#### 3.1.1 FUNCTION PROPERTIES FROM AXIOMATIC DEFINITION OF COUNTERFACTUALS

To determine the properties of such functions, we review Pearl's axiomatic definition of counterfactuals (Pearl, 2009, Sec. 7.3.1). The *soundness* theorem states that the properties of composition, effectiveness and reversibility hold true in all causal models (Galles & Pearl, 1998). The *completeness* theorem states that these properties are complete (Halpern, 1998). Together these theorems state that composition, effectiveness and reversibility are the necessary and sufficient properties of counterfactuals in any causal model. We aim to construct a functional system that obeys these axioms.

**Effectiveness:** Intervening on a variable to have a specific value will cause the variable to take on that value. Thus, suppose $\mathrm{Pa}(\cdot)$ is an oracle function that returns the parents of a variable, then we have the following equality: $\mathrm{Pa}(\mathrm{f}(x, \mathbf{pa}, \mathbf{pa}^*)) = \mathbf{pa}^*$.

**Composition:** Intervening on a variable to have the value it would otherwise have without the intervention will not affect other variables in the system. This implies the existence of a null transformation $\mathrm{f}(x, \mathbf{pa}, \mathbf{pa}) = x$ since if $\mathbf{pa}^* = \mathbf{pa}$, then $x$ is not affected.

**Reversibility:** Reversibility prevents the existence of multiple solutions due to feedback loops. In recursive systems such as DAGs, it follows trivially from composition. However, in the general case, these properties are independent. If setting a variable $A$ to a value $a$ results in a value $b$ for a variable $B$, and setting $B$ to a value $b$ results in a value $a$ for $A$, then $A$ and $B$ will take the values $a$ and $b$. If a mechanism is invertible, taking a twin network conception of the SCM (see Pearl, 2009, Sec. 7.1.4) and replacing the previous variables with the observation $x$ and its counterfactual $x^*$, it follows that if $x^* \coloneqq \mathrm{f}(x, \mathbf{pa}, \mathbf{pa}^*)$, then $x = \mathrm{f}(x^*, \mathbf{pa}^*, \mathbf{pa})$. In other words, the mapping between the observation and the counterfactual is deterministic for invertible mechanisms. Otherwise, there would be a feedback loop. See Appendix A for proof.

#### 3.1.2 PARTIAL COUNTERFACTUAL FUNCTIONS

We can also consider partial counterfactual functions, which model the effect of a single parent on the observation *independently* of all other causes. These partial interventions allow us to decompose an intervention into steps whereby a single parent is changed while holding the remaining parents fixed. Each step acts on all parents, but only one parent changes value at a time: $x^* = f_k(x, pa_k, pa_k^*)$, where $\mathbf{pa}^* = \mathbf{pa}_{\mathcal{K}\setminus k} \cup \{pa_k^*\}$. Notice that this is still an intervention on all parents, in contrast with atomic interventions where only one variable is intervened upon and all others updated according to the SCM. These partial functions can be obtained by taking a full counterfactual function and fixing all parents to their initial values except one: $f_k(x, pa_k, pa_k^*) = f\Big(x, \mathbf{pa}_{\mathcal{K}\setminus k} \cup \{pa_k\}, \mathbf{pa}_{\mathcal{K}\setminus k} \cup \{pa_k^*\}\Big)$, or they can directly implicitly infer the values of the fixed parents, as shown in Appendix C. In Appendix C.1, we show that a full intervention can be decomposed into a sequence of partial interventions and that the partial functions must obey the commutative property.

### 3.2 Measuring soundness of counterfactuals

We now distinguish between the ideal counterfactual function $f(\cdot)$ and its approximation $\hat{f}(\cdot)$. We consider a scenario where we want to evaluate how good our estimate of a counterfactual function is from observational data alone, without access to observed counterfactuals. Using the axiomatic properties described in Section 3.1.1, we can derive a set of soundness metrics that compare the approximate and ideal models.

**Composition:** Since the ideal model cannot change an observation under the null transformation, we can measure how much the approximate model deviates from the ideal by calculating the distance between the original observation and the $m$th time null-transformed observation. The repeated application of the function will highlight what types of corruptions the approximate model produces on the observation. Given a distance metric $d_X(\cdot, \cdot)$, such as the $l_1$ distance, an observation $x$ with parents $\mathbf{pa}$ and a functional power $m$, we can measure composition as:

$$\text{composition}^{(m)}(x, \mathbf{pa}) := d_X\Big(x, \hat{f}^{(m)}(x, \mathbf{pa}, \mathbf{pa})\Big). \tag{1}$$

See Appendix B for a discussion on desirable metric properties.

**Reversibility:** When a mechanism is invertible, the ideal model must be cycle-consistent. Thus we can measure reversibility by calculating the distance between the original observation and the $m$th time cycled-back transformed observation. Setting $\hat{p}(x, \mathbf{pa}, \mathbf{pa}^*) := \hat{f}\Big(\hat{f}(x, \mathbf{pa}, \mathbf{pa}^*), \mathbf{pa}^*, \mathbf{pa}\Big)$. Given a distance metric $d_X(\cdot, \cdot)$, an observation $x$ with parents $\mathbf{pa}$ and a functional power $m$, we can measure reversibility as:

$$\text{reversibility}^{(m)}(x, \mathbf{pa}, \mathbf{pa}^*) := d_X\Big(x, \hat{p}^{(m)}(x, \mathbf{pa}, \mathbf{pa}^*)\Big). \tag{2}$$

Note that in most real-world scenarios, the inherent uncertainty regarding exogenous factors makes it hard to determine whether the true mechanism is invertible.

**Effectiveness:** Unlike composition and reversibility, which we can measure independently of the data distribution, effectiveness is difficult to measure objectively without relying on data-driven methods or strong domain knowledge. We propose measuring effectiveness individually for each parent by creating a pseudo-oracle function $\widehat{Pa_k}(\cdot)$, which returns the value of the parent $pa_k$ given the observation. These functions can be human-made programs or machine learning models learnt from data via classification/regression. The inevitable limitation of the data-driven approach is the approximation error. Moreover, we must be especially cautious in the presence of confounded parents to ensure that the pseudo-oracles do not exploit shortcuts and correctly retrieve the desired parent. To independently measure how well the effect of each parent is modelled, we measure effectiveness after applying partial counterfactual functions (see Section 3.1.2). Using an appropriate distance metric $d_k(\cdot, \cdot)$, such as accuracy for discrete variables or $l_1$ distance for continuous ones, we measure effectiveness for each parent as:

$$\text{effectiveness}_k(x, \mathbf{pa}, \mathbf{pa}^*) = d_k\Big(\widehat{Pa_k}\big(\hat{f}_k(x, pa_k, pa_k^*)\big), pa_k^*\Big). \tag{3}$$

### 3.3 Simulated Interventions

When learning deep models from biased data, we must be careful not to allow the model to learn shortcuts which do not reflect the true causal relationships in the data. While some generative models have inductive priors that make them more robust to confounding (Higgins et al., 2017; Li et al., 2021), discriminative models are quite brittle (Geirhos et al., 2020). The causal approach to address confounding is to break the offending causal links in the SCM via an intervention, removing the possibility of learning shortcuts by de-biasing the data.

Consider the joint distribution of the data $P(x, \mathbf{pa}_{\mathcal{K}\setminus k}, \mathrm{pa}_k) = P(x|\mathrm{pa}_k, \mathbf{pa}_{\mathcal{K}\setminus k}) P(\mathrm{pa}_k, \mathbf{pa}_{\mathcal{K}\setminus k})$, where $\mathcal{K}$ is the set of all parents of $x$. If we perform a soft intervention (see Peters et al., 2017, Sec. 3.2) where we set $\mathrm{pa}_k$ and $\mathbf{pa}_{\mathcal{K}\setminus k}$ to their respective marginal distributions, we obtain an interventional distribution where $\mathrm{pa}_k$ and $\mathbf{pa}_{\mathcal{K}\setminus k}$ are independent. The joint distribution now factorises as:

$$P^{\mathrm{do}(\mathrm{pa}_k \sim P(\mathrm{pa}_k); \mathbf{pa}_{\mathcal{K}\setminus k} \sim P(\mathbf{pa}_{\mathcal{K}\setminus k}))}(x, \mathbf{pa}_{\mathcal{K}\setminus k}, \mathrm{pa}_k) = P(x|\mathbf{pa}_{\mathcal{K}\setminus k}, \mathrm{pa}_k) P(\mathrm{pa}_k) P(\mathbf{pa}_{\mathcal{K}\setminus k}) \quad (4)$$

In the absence of interventional data, we use simulated interventions by resampling the data – note that this is only possible when the observed joint distribution has full support over the product of marginals. We sample $\mathrm{pa}_k$ and $\mathbf{pa}_{\mathcal{K}\setminus k}$ according to their respective marginal distributions and then randomly sample an observation conditioned on the sampled parent values. To make all parents independent of each other, following the same logic, we can sample each parent independently according to its marginal distribution.

### 3.4 Deep generative models as approximate counterfactual functions

In this section, we discuss two generative models commonly used as approximate counterfactual inference engines.

**Conditional VAE:** The evidence lower bound for a classic conditional VAE (Kingma & Welling, 2014; Higgins et al., 2017) with a $\beta$ penalty is given by:

$$\mathrm{ELBO}_\beta(\theta, \omega) = \mathbb{E}_{q_\theta(z|x,\mathbf{pa})}[\log p_\omega(x|z, \mathbf{pa})] - \beta\, D_{\mathrm{KL}}[q_\theta(z|x, \mathbf{pa}) \| p(z)], \quad (5)$$

where $q_\theta(z|x, \mathbf{pa})$ is a normal distribution parameterised by a neural network encoder with parameters $\theta$, $p_\omega(x|z, \mathbf{pa})$ is a set of pixel-wise independent Bernoulli or Normal distributions parameterised by a neural network decoder with parameters $\omega$, and $p(z)$ is an isotropic normal prior distribution. The model is trained by maximising the ELBO with respect to parameters of the neural networks using the re-parametrisation trick to sample from the approximate latent posterior: $z = \mu_\theta(x, \mathbf{pa}) + \sigma_\theta(x, \mathbf{pa}) \odot \epsilon_z$ where $\epsilon_z \sim \mathcal{N}(0, I)$. The construction of the conditional VAE naturally leads to a composition constraint via the likelihood term. Even though the model can ignore the conditioning, using a bottleneck or a $\beta$ penalty pushes the model towards using the conditioning, thus enforcing effectiveness, making the conditional VAE a natural choice for a counterfactual model.

We can produce counterfactuals by encoding an observation and its parents, sampling the latent posterior, and then decoding it along with the counterfactual parents: $x^* \sim p_\omega(x|z, \mathbf{pa}^*)$, where $z \sim q_\theta(z|x, \mathbf{pa})$. Rewriting the previous expression as: $x^* := \hat{\mathrm{f}}_{\theta,\omega}(x, \mathbf{pa}, \mathbf{pa}^*)$ where $\hat{\mathrm{f}} \sim P(\hat{\mathrm{f}})$, we see the parallels to the formulation in Section 3.1. Even though it is possible to generate new samples from the model by sampling $z$, the VAE is not a full causal generative model. The latent variable $z$ is not the same as the exogenous noise of the SCM $\epsilon$. There are no guarantees that $z \perp\!\!\!\perp \mathrm{pa}|x$ or that the exogenous noise would be normally distributed. Furthermore, there are no guarantees that the forward model (decoder) can disentangle the effects of each parent on the observation.

**Conditional GAN with a composition constraint:** Given a joint distribution $x, \mathbf{pa} \sim P(x, \mathbf{pa})$ and the marginal distribution of each parent $\mathrm{pa}_k \sim P(\mathrm{pa}_k)$, if we independently sample each parent according to its marginal and perform an intervention, we obtain an interventional distribution $x, \mathbf{pa} \sim P^{do}(x, \mathbf{pa})$, where the parents are independent of each other $\mathbf{pa} \sim \prod_k P(\mathrm{pa}_k)$. We can obtain this distribution via a simulated intervention or by applying a counterfactual function to samples of a source distribution $P^{src}(x, \mathbf{pa})$, which can be the joint or the interventional distribution itself. Since these distributions must be equal, we can use GANs to minimise statistical divergence between the two (Goodfellow et al., 2014; Nowozin et al., 2016):

$$F(\theta, \omega) = \mathop{\mathbb{E}}_{x,\mathbf{pa}\sim P^{do}(x,\mathbf{pa})}\left[\log\left(D_\omega(x, \mathbf{pa})\right)\right] - \mathop{\mathbb{E}}_{\substack{x,\mathbf{pa}\sim P^{src}(x,\mathbf{pa}) \\ \mathrm{pa}_k^* \sim P(\mathrm{pa}_k)}}\left[\log\left(1 - D_\omega\bigl(\hat{\mathrm{f}}_\theta(x, \mathbf{pa}, \mathbf{pa}^*), \mathbf{pa}\bigr)\right)\right],$$
(6)

where the conditional generator $\hat{\mathrm{f}}_\theta(x, \mathbf{pa}, \mathbf{pa}^*)$ is a neural network parameterised by $\theta$ which approximates the counterfactual function, $D_\omega$ is the critic function parameterised by parameters $\omega$, and $F(\theta, \omega)$ is minimised with respect to $\theta$ and maximised with respect to $\omega$.

Unlike VAEs, GANs have no inherent mechanism to enforce composition; thus, we introduce a composition constraint which encourages the null transformed observation to be close to the original observation. Given distance metric $d(\cdot, \cdot)$ such as the $l_2$ distance, we can write the following regulariser, which we add to the GAN objective:

$$R_{\text{composition}}(\theta) = \mathop{\mathbb{E}}_{x,\mathbf{pa} \sim P^{src}(x,\mathbf{pa})} d\Big(x, \hat{f}_\theta(x, \mathbf{pa}, \mathbf{pa})\Big). \qquad (7)$$

Similarly, for invertible mechanisms, we can add a reversibility constraint. However, early experiments proved it redundant while adding computational cost. The proposed GAN model has no distribution over functions, which is equivalent to assuming that the exogenous noise posterior $P(\epsilon|x, \mathbf{pa})$ is delta distributed. Although this is a strong assumption, assuming a normal posterior as is done in the VAE is also restrictive. Both are unlikely to be a good approximation of the true noise distribution in most real-world scenarios.

## 4 EXPERIMENTS AND RESULTS

We now demonstrate the utility of our evaluation framework by applying it to three datasets. For demonstration purposes, we assume invertible mechanisms so we can use the reversibility metric.

### 4.1 COLOUR MNIST

To illustrate the effect confounding has on counterfactual approximation, we construct a simple experiment using the MNIST dataset (LeCun et al., 1998) where we introduce a new parent: the digit's hue. We colour each image by triplicating the grey-scale channel, setting the saturation to 1 and setting the hue to a value between 0 and 1. The hue value is given by one of three possible SCMs:

- **Unconfounded:** where we draw the hue from a uniform distribution independently of the digit: hue $\sim \text{Uniform}(0, 1)$;
- **Confounded without full support:** where the hue depends on the digit but the joint distribution does not have full support: hue $\sim \mathcal{N}(\text{digit}/10 + 0.05, \sigma)$, where $\sigma = 0.05$;
- **Confounded with full support:** like the previous scenario except we include a percentage of outliers where the hue is drawn independently of the digit in order to ensure the distribution has full support: $b \sim \text{Bernoulli}(p)$ and hue $\sim \mathcal{N}(\text{digit}/10 + 0.05, \sigma)$ if $b = 0$ else hue $\sim \text{Uniform}(0, 1)$, where $\sigma = 0.05$ and $p = 0.01$.

Figure 4 shows the joint distribution of digit and hue for the three SCMs, and Figure 5 shows samples from the confounded and unconfounded joint distributions.

We compare the counterfactual soundness of the following models: a VAE with Bernoulli log-likelihood and $\beta = \{1, 2\}$, a VAE with Normal log-likelihood with a fixed variance of 0.1 and $\beta = 5$, and a conditional GAN with a composition constraint. Model and training details can be found in Appendix D.2. To obtain lower bounds for the soundness metrics, we include two models that, by design, cannot perform abduction: the identity function and a VAE without the encoder at inference time (Bernoulli VAE, $\beta = 1$). For the confounded scenarios, we perform a simulated intervention to break the causal link between digit and hue. Since the hue is a continuous variable, to simulate the intervention, we calculate the histogram of its marginal distribution (5 bins) and resample it as if it were a discrete variable. We train the models on data generated from the three SCMs but always test on the unconfounded test set, which mimics a scenario where the correlations in the training data are spurious and ensures the results of the tests are not biased. Note that if the test set is biased, we can obtain biased estimates for the soundness metrics; this setting falls out of the scope of this study. We use accuracy for the digit and the absolute error for the hue. To measure composition and reversibility, we use the $l_1$ distance. Table 1 shows the results for all models and scenarios. For the VAEs, we sample a unique function for each observation but keep it fixed for repeated interventions.

**Comparing with poor counterfactuals.** We can see that the identity function achieves perfect composition and reversibility but, as expected, fails at effectiveness. In contrast, the VAE without encoder performs well in terms of effectiveness but fails at composition and reversibility. Interestingly, the same VAE with the encoder achieves a composition after ten null interventions of $l_1^{(10)} = 17.36$ (0.62), close to the value without the encoder $l_1^{(10)} = 19.04$ (0.09), indicating that the VAE progressively loses the identity of the image and converges to a random sample (Figure 1a).

Table 1: Soundness metrics on colour MNIST over 5 random seeds. We measure composition after the null intervention and reversibility after one intervention cycle. We measure effectiveness using digit accuracy and hue absolute error in percentage points since hue $\in [0, 1]$. * GAN always requires simulated intervention for target distribution and thus cannot be trained w/o full support.

| dataset | inter-<br>ven-<br>tion | model | null-intervention<br>composition<br>$l_1^{(1)} \downarrow$ | digit intervention | | | hue intervention | | |
|---|---|---|---|---|---|---|---|---|---|
| | | | | effectiveness | | reversibility | effectiveness | | reversibility |
| | | | | $\mathrm{acc}_{\mathrm{digit}}(\%) \uparrow$ | $\mathrm{ae}_{\mathrm{hue}}(\%) \downarrow$ | $l_1^{(1)} \downarrow$ | $\mathrm{acc}_{\mathrm{digit}}(\%) \uparrow$ | $\mathrm{ae}_{\mathrm{hue}}(\%) \downarrow$ | $l_1^{(1)} \downarrow$ |
| un-<br>con-<br>found-<br>ed | - | Identity | 0.00 | 10.50 | 1.38 | 0.00 | 99.18 | 32.98 | 0.00 |
| | - | VAE w/o encoder | 19.04 (0.09) | 97.08 (0.25) | 1.32 (0.05) | 19.04 (0.09) | 97.24 (0.26) | 1.32 (0.06) | 19.04 (0.09) |
| | - | Bernoulli VAE $\beta$=1 | 5.98 (0.06) | 98.68 (0.13) | 1.29 (0.04) | 7.67 (0.06) | 99.45 (0.09) | 1.26 (0.05) | 7.24 (0.05) |
| | - | Bernoulli VAE $\beta$=2 | 6.86 (0.07) | 99.52 (0.07) | 1.33 (0.15) | 9.10 (0.12) | 99.60 (0.04) | 1.32 (0.15) | 8.62 (0.11) |
| | - | Normal VAE $\beta$=5 | 6.26 (0.29) | 97.24 (0.26) | 1.52 (0.28) | 8.07 (0.26) | 99.38 (0.06) | 1.47 (0.27) | 7.51 (0.32) |
| | - | GAN | 4.92 (0.05) | 94.28 (1.01) | 1.60 (0.22) | 9.22 (0.27) | 98.98 (0.05) | 1.55 (0.23) | 5.60 (0.03) |
| con-<br>found-<br>ed<br>w/o<br>full<br>support | no | Bernoulli VAE $\beta$=1 | 9.20 (1.31) | 97.12 (1.05) | 10.74 (4.77) | 11.42 (1.49) | 98.89 (0.16) | 11.60 (6.14) | 11.11 (1.61) |
| | no | Bernoulli VAE $\beta$=2 | 10.84 (0.45) | 98.94 (0.17) | 10.36 (1.39) | 12.82 (0.45) | 99.17 (0.05) | 10.07 (1.39) | 12.52 (0.41) |
| | no | Normal VAE $\beta$=5 | 11.21 (0.63) | 94.74 (0.51) | 14.17 (2.63) | 13.32 (0.62) | 98.81 (0.22) | 14.27 (3.03) | 12.69 (0.59) |
| | yes | Bernoulli VAE $\beta$=1 | 8.63 (0.50) | 96.94 (0.26) | 6.38 (1.58) | 11.10 (0.75) | 98.88 (0.25) | 7.02 (1.96) | 10.79 (0.75) |
| | yes | Bernoulli VAE $\beta$=2 | 9.85 (0.33) | 95.76 (1.63) | 6.44 (1.24) | 12.10 (0.39) | 95.77 (1.56) | 6.44 (1.37) | 11.86 (0.29) |
| | yes | Normal VAE $\beta$=5 | 9.32 (1.41) | 95.35 (0.71) | 7.54 (1.99) | 11.29 (1.39) | 98.79 (0.28) | 7.30 (2.03) | 10.85 (1.36) |
| con-<br>found-<br>ed<br>w/<br>full<br>support | no | Bernoulli VAE $\beta$=1 | 6.68 (0.27) | 96.62 (2.09) | 8.52 (6.93) | 8.89 (0.70) | 99.20 (0.10) | 12.15 (11.69) | 8.45 (0.69) |
| | no | Bernoulli VAE $\beta$=2 | 7.56 (0.10) | 99.36 (0.16) | 2.70 (0.12) | 9.67 (0.06) | 99.47 (0.05) | 2.54 (0.12) | 9.32 (0.09) |
| | no | Normal VAE $\beta$=5 | 6.72 (0.30) | 95.53 (0.28) | 3.88 (1.12) | 9.06 (0.68) | 99.07 (0.04) | 3.59 (1.20) | 8.45 (0.67) |
| | yes* | GAN | 6.05 (0.06) | 95.17 (0.55) | 1.95 (0.07) | 11.18 (0.10) | 99.18 (0.10) | 1.73 (0.11) | 7.79 (0.10) |
| | yes | Bernoulli VAE $\beta$=1 | 6.67 (0.10) | 99.07 (0.15) | 2.31 (0.24) | 8.42 (0.16) | 99.37 (0.13) | 3.08 (1.08) | 8.40 (0.48) |
| | yes | Bernoulli VAE $\beta$=2 | 7.84 (0.09) | 99.63 (0.03) | 2.16 (0.06) | 9.63 (0.08) | 99.61 (0.06) | 2.01 (0.10) | 9.34 (0.09) |
| | yes | Normal VAE $\beta$=5 | 6.51 (0.29) | 97.75 (0.18) | 3.05 (0.44) | 8.35 (0.29) | 99.31 (0.07) | 2.73 (0.47) | 7.83 (0.31) |
| | yes | GAN | 5.25 (0.06) | 96.27 (0.26) | 1.84 (0.11) | 10.75 (0.34) | 99.01 (0.06) | 1.77 (0.14) | 6.20 (0.04) |

**Comparing models and scenarios.** For each scenario, we can see how the proposed metrics allow us to compare models independently of the model class and directly compare the quality of counterfactual approximations. For the confounded scenario without full support, we see a significant drop in performance which we cannot recover even when using a simulated intervention since the lack of support prevents us from sampling points from some areas of the joint distribution. Nevertheless, when the confounded distribution has full support, we can recover the performance using a simulated intervention or selecting an appropriate $\beta$ penalty for the VAE. In addition to the numerical results, it is also helpful to visualise the results of the tests. To highlight the impact of confounding, in Figure 1, we show the soundness tests for a disentangled model (Normal VAE on the confounded scenario w/ full support and a simulated intervention) and an entangled model (Normal VAE on the confounded scenario w/o full support and no simulated intervention). We can see that the entangled model cannot preserve the image's identity, quickly changing its colour and distorting its shape (Figure 1d). In contrast, the disentangled model is far more capable of preserving the image's identity even if the shape gets distorted over repeated applications (Figure 1a). Regarding effectiveness, we can see that the entangled model consistently fails to change the digit without changing its hue (Figure 1e). Conversely, the disentangled model can change the digit independently of the hue, and its counterfactuals seem to preserve properties such as slant and thickness. Finally, regarding hue reversibility, we can see that only the disentangled model is able to correctly cycle back and forth between hue values even if there is some shape distortion (Figures 1f and 1c).

**Measuring effectiveness using pseudo-oracles trained on biased data.** In the previous experiment, we used pseudo-oracles trained on unconfounded data to compare models. However, in real-world scenarios where we do not control the data-generation process, we have to use pseudo-oracles trained on the available data. If we use pseudo-oracles trained on biased data, we can get biased estimates for effectiveness. We can use simulated interventions when training the oracles to address this problem. When training the pseudo-oracles on the confounded scenario w/ full support, we obtain a digit accuracy of 90.67% and hue absolute error of 6.34% – down from 99.18% and 1.38% respectively, for the oracles trained on unconfounded data. However, using a simulated intervention, we obtained 96.70% and 4.11% for the digit accuracy and hue absolute error, respectively. Regardless, we show in Tables 2, 3 that the relative performance ranking of the models is only slightly affected by the decrease in oracle quality due to confounding. Additionally, we show in Table 4 that the ranking is also robust to using linear models (logistic/linear regression) as pseudo-oracles, giving us some margin for error when measuring the relative effectiveness.

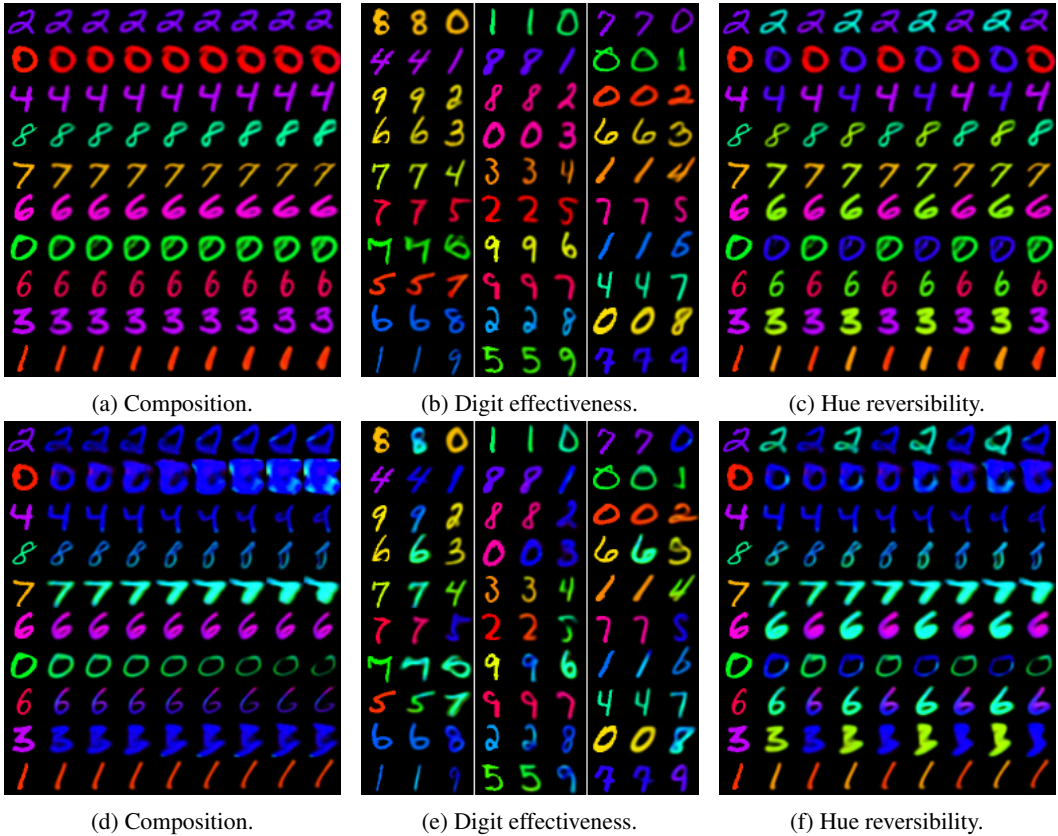

Figure 1: Soundness tests on colour MNIST for a disentangled model **(a, b, c)** and a entangled model **(d, e, f)**. **Composition (a, d)**: one sample per row, the first column is the input, each column after is the result of applying the null intervention to the previous column. **Digit effectiveness (b, e)**: three sub-panels with one sample per row, the first column is the input, the second column is the result of the null intervention, the last column is the result of a partial intervention on the digit. **Hue reversibility (c, f)**; one sample per row, the first column is the input, the second column is the result of a partial intervention on the hue, the third column is the result of applying the partial intervention that cycles back to the original hue. The remaining columns repeat the cycle.

### 4.2 3D Shapes

Next, we test our evaluation framework on the 3D shapes dataset (Burgess & Kim, 2018), which comprises procedurally generated images from 6 independent parents: floor colour, wall colour, object colour, scale, shape and orientation. Each possible combination of attributes results in an image creating a dataset of 480000 images. We keep 10% of images as a test set and train on the remaining 90%. We treat all variables as discrete since even the continuous ones, such as colour, only take a set of discrete values. We compare the counterfactual properties of a Bernoulli VAE ($\beta = 1$) with a constrained GAN. Model and training details are available in Appendix E.1. Figures 2 (a-d) and 2 show the composition/effectiveness plots for both models. The models can successfully change all parents without perturbing the remainder (Table 5). However, we can see that the GAN introduces a slight distortion in the image, which the VAE does not. In fact, the VAE achieves a composition score of $l_1^{(1)} = 1.62 \pm (0.07)$ and $l_1^{(10)} = 1.62(0.07)$ while the GAN has a score of $l_1^{(1)} = 5.11(0.14)$ and $l_1^{(10)} = 5.16(0.15)$. The lack of exogenous factors explains the consistency in the composition scores after repeated null interventions. Abduction is trivial since there is nothing to abduct. Thus, the model can ignore the observation and rely solely on the parents to generate the counterfactual.

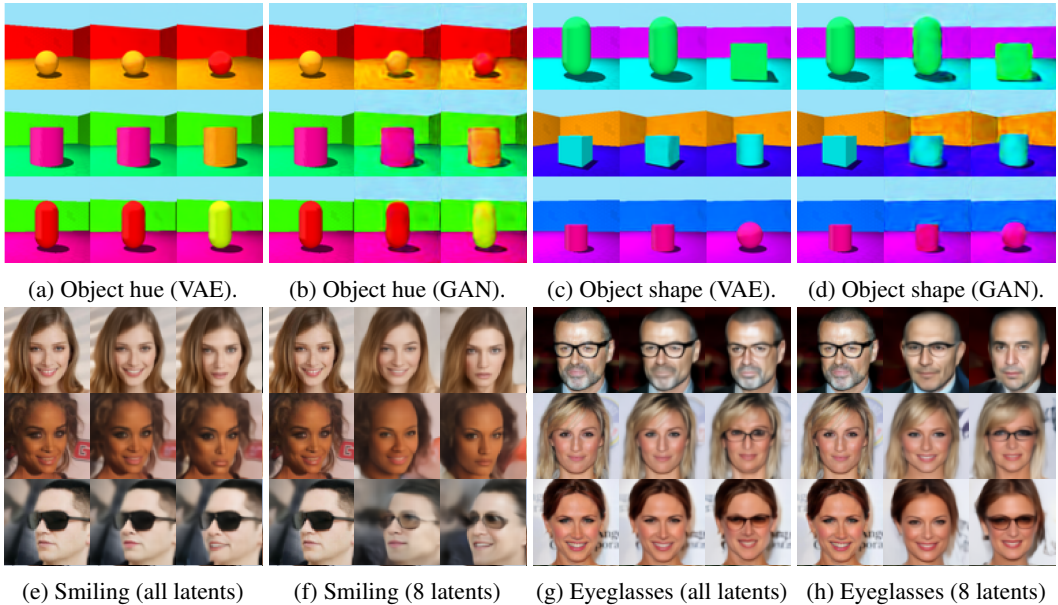

(a) Object hue (VAE). (b) Object hue (GAN). (c) Object shape (VAE). (d) Object shape (GAN).

(e) Smiling (all latents) (f) Smiling (8 latents) (g) Eyeglasses (all latents) (h) Eyeglasses (8 latents)

Figure 2: Effectiveness/Composition on 3D shapes (a-d) and CelebA-HQ (e-h). Each sub-figure has one sample per row: the first column is the input, the second column is the result of the null intervention, and the last column is the result of a partial intervention on a specific parent.

## 4.3 CELEBA-HQ

Lastly, we demonstrate our evaluation framework on a natural image dataset which requires more complex generative models. We used the CelebA-HQ dataset (Karras et al., 2018) with a 64x64 resolution and the binary parent attributes 'smiling' and 'eyeglasses'. For the model, we opted to extend the VDVAE (Child, 2021) to a conditional model to enable counterfactuals (see Appendix F). Since the latent code is hierarchical, the abduction process can be decomposed into abducting subsets of latent variables at different resolutions. This added complexity enables higher fidelity images but makes abduction harder in practice. When producing counterfactuals, abducting all latent variables can result in the model ignoring the counterfactual parent conditioning. Conversely, abducting only a subset trends the model towards obeying the conditioning at the cost of faithfulness to the input observation. In other words, there is a trade-off between effectiveness and composition/reversibility. Figures 2e and 2g show 'smiling' and 'eyeglasses' counterfactuals for a model with 42 latent variables when abducting all variables. Figures 2f and 2h show the same counterfactuals when abducting only a subset of 8 variables. We can see that, under full abduction, it is harder for the counterfactuals to obey the conditioning, but they are more faithful to the input (e.g. higher 'smiling' composition $l_1^{(1)} = 3.657 \pm 0.0006$ but lower effectiveness F1-score $= 0.848 \pm 0.0006$. In contrast, when using partial abduction, we obtain higher effectiveness F1-score $= 0.933 \pm 0.002$ at the cost of much lower composition $l_1^{(1)} = 20.890 \pm 0.018$. Table 6 and Figure 7 show the evolution of effectiveness, composition, and reversibility with the number of fixed posterior latent variables.

## 5 DISCUSSION

We have presented a theoretically grounded framework for evaluating approximate counterfactual inference models without observed counterfactuals or knowledge of the underlying SCM. While guarantees of truly identifiable models using deep learning are not possible in the general case, we show that we can measure the *soundness* of such models by observing which constraints they should obey in the ideal case. Further, we show the impact of confounding on training and evaluation of deep counterfactual inference models and how its effect can be mitigated by using simulated interventions. We hope the ideas presented here can help inform the development and evaluation of future approximate counterfactual inference models that use deep models as their base.

ETHICS STATEMENT

The ability to generate plausible image counterfactuals can have both productive and nefarious applications. On the positive side, counterfactual explanations have the potential to improve the interpretability of deep learning models and help bridge the gap between human and machine intelligence. Counterfactual queries may help to identify disparities in model performance, and counterfactual data augmentation can mitigate dataset bias against underrepresented groups in downstream tasks such as classification. With that said, if incorrectly used, counterfactual image models may also further exacerbate such biases. Moreover, visually plausible artificially generated counterfactual images could be misused by ill-intended parties to deceive, mislead or spread misinformation. We argue that the opportunities and risks of counterfactual image generation must be carefully considered throughout development and a comprehensive evaluation framework is integral to this process.

ACKNOWLEDGEMENTS

This project has received funding from the ERC under the EU's Horizon 2020 research and innovation programme (grant No. 757173). We thank Dimitrios Vytiniotis for the helpful discussion.

## A  REVERSIBILITY PROOF

**Lemma A.1.** *Let $x = g(\epsilon, \mathbf{pa})$ be a mechanism where $x$ is the observation with parents $\mathbf{pa}$ and exogenous noise variable $\epsilon$. Further, let $x^*$ be a counterfactual of $x$ with parents $\mathbf{pa}^*$. If the mechanism is invertible, then the exogenous noise is deterministically given by: $\epsilon = g^{-1}(x, \mathbf{pa})$ or $\epsilon = g^{-1}(x^*, \mathbf{pa}^*)$. Under these conditions, in the counterfactual function form we have that if $x^* := f(x, \mathbf{pa}, \mathbf{pa}^*)$, then $x = f(x^*, \mathbf{pa}^*, \mathbf{pa})$.*

*Proof.*

$$\begin{aligned} x^* &= g(\epsilon, \mathbf{pa}^*) \\ &= g\big(g^{-1}(x, \mathbf{pa}), \mathbf{pa}^*\big) \\ &= f(x, \mathbf{pa}, \mathbf{pa}^*) \end{aligned} \tag{8}$$

And:

$$\begin{aligned} x &= g(\epsilon, \mathbf{pa}) \\ &= g\big(g^{-1}(x^*, \mathbf{pa}^*), \mathbf{pa}\big) \\ &= f(x, \mathbf{pa}^*, \mathbf{pa}) \end{aligned} \tag{9}$$

□

## B  DISTANCE METRICS

Given two points $a \in \mathbb{R}^N$ and $b \in \mathbb{R}^N$ and a distance $d_X(\cdot, \cdot) : (\mathbb{R}^N, \mathbb{R}^N) \to \mathbb{R}^+$, for the distance to be considered a metric, it must obey the following properties (see Phillips, 2021, Sec. 6.1):

- Non-negativity: $d_X(a, b) \geq 0$;
- Identity: $d_X(a, b) = 0$ if $a = b$;
- Symmetry: $d_X(a, b) = d_X(b, a)$;
- Triangular inequality: $d_X(a, b) \leq d_X(a, c) + d_X(c, b)$.

In this paper, we opted to use the $l_1$ distance, which is a metric, because we can directly interpret its value as the average pixel intensity by which two images differ. While there is an argument to be made for the use of perceptual distances for images, they are not metrics and have the same data-driven weaknesses, which pose a problem for our pseudo-oracles. As a result, we opted for a data-independent distance/metric.

## C  PARTIAL COUNTERFACTUAL FUNCTIONS

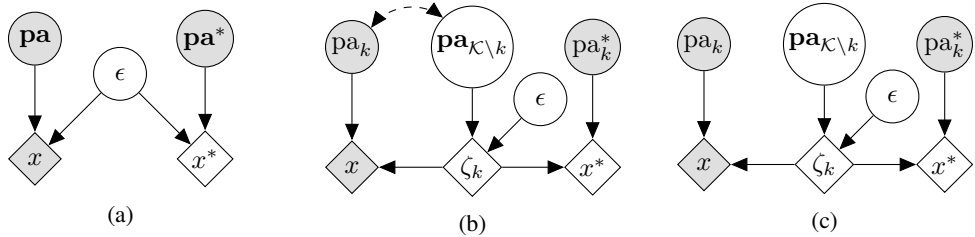

Figure 3: Process of making one parent independent of the rest using a simulated intervention. (a) Starting SCM; (b) SCM after separating the target parent from the rest; (c) SCM after the intervention.

Figure 3a shows a causal graph for a single mechanism, where there can exist unknown arbitrary causal links between the variable's parents. Suppose we take the endogenous causes of $x$ and convert all but one into exogenous causes resulting in the causal graph in Figure 3b. The bidirectional arrow between the endogenous parent $\mathrm{pa}_k$ and the now exogenous parents $\mathbf{pa}_{\mathcal{K} \setminus k}$ denotes the unknown causal relationships between them. This manipulation results in a "partial" mechanism $x^* := g_k(\zeta_k, \mathrm{pa}_k)$ where only one parent is endogenous and thus susceptible to action. The remaining parents and the previous exogenous noise are grouped into a new noise variable $\zeta_k := g_k^{-1}(x, \text{pa}_k)$. However, due to the dependence between the target parent $\text{pa}_k$ and the unobserved exogenous parents $\mathbf{pa}_{\mathcal{K}\setminus k}$, abducting $\zeta_k$ is impossible. The path between $\zeta_k$ and $\text{pa}_k$ breaks the assumption of independent exogenous noise variables, making it impossible to separate the effects of endogenous and exogenous causes. The mechanism induced by $\zeta_k$ now depends on the cause $\text{pa}_k$ since $\zeta_k \not\perp\!\!\!\perp \text{pa}_k$, violating the principle of independent causal mechanisms.

To preserve the principle of independent causal mechanisms, we must first make $\text{pa}_k$ independent of the remaining parents $\mathbf{pa}_{\mathcal{K}\setminus k}$. Since the causal direction between $\text{pa}_k$ and $\mathbf{pa}_{\mathcal{K}\setminus k}$ is unknown, we intervene on both these variables, thus guaranteeing their link is severed. The modified SCM is shown in Figure 3c, where abduction is now possible since the mechanism no longer depends on the cause. Because there are no backdoor paths between $\text{pa}_k$ and $\zeta_k$, using *do*-calculus (Pearl, 2009), we have that $P(\zeta_k|x, \text{do}(\text{pa}_k)) = P(\zeta_k|x, \text{pa}_k)$. Since $x^*$ depends on $x$ via $\zeta_k$ the distribution over potential counterfactuals $P(x^*|x, \text{pa}_k)$ remains unchanged after the intervention. After the intervention, we can learn the partial counterfactual function without confounding being an issue. The partial function changes one parent while holding the remainder fixed since these are now part of the exogenous noise.

### C.1 PARTIAL COUNTERFACTUAL FUNCTION DECOMPOSITION

Using the process described in Section 3.1, we can write a partial counterfactual function as: $x^* := f_k(x, \text{pa}_k, \text{pa}_k^*)$ where $f_k \sim P(f_k)$. Since the output of $f_k(\cdot)$ is counterfactual of $x$, all the properties from Section 3.1.1 still apply. Repeating the process for each parent results in a set of independent partial functions allowing us to write the full counterfactual function as a composition of multiple independent partial functions:

$$x^* := \big[f_K(\cdot, \text{pa}_K, \text{pa}_K^*) \circ \ldots \circ f_2(\cdot, \text{pa}_2, \text{pa}_2^*) \circ f_1(\cdot, \text{pa}_1, \text{pa}_1^*)\big](x) \quad (10)$$

By construction, each partial function models the effect of a single parent on the observation independently of other parents. As a result, the decomposition in equation 10 must be commutative.

**Lemma C.1.** *Let $x = g(\epsilon, \text{pa}_1, \text{pa}_2)$ be the generative model for an observation $x$ with parents $\text{pa}_1$ and $\text{pa}_2$ and exogenous noise variable $\epsilon$. Further, let $f_1(x, \text{pa}_1, \text{pa}_1^*)$ and $f_2(x, \text{pa}_2, \text{pa}_2^*)$ be two partial functions operating on $\text{pa}_1$ and $\text{pa}_2$ respectively. Then, the composition of $f_1(\cdot)$ and $f_2(\cdot)$ is commutative:*

$$f_1(f_2(x, \text{pa}_2, \text{pa}_2^*), \text{pa}_1, \text{pa}_1^*) = f_2(f_1(x, \text{pa}_1, \text{pa}_1^*), \text{pa}_2, \text{pa}_2^*) \quad (11)$$

*Proof.* Since any counterfactual of $x$ can be generated by the generative model $g(\cdot)$, we have:

$$\begin{aligned} f_1(x, \text{pa}_1, \text{pa}_1^*) &= g(\epsilon, \text{pa}_1^*, \text{pa}_2) \\ f_2(x, \text{pa}_2, \text{pa}_2^*) &= g(\epsilon, \text{pa}_1, \text{pa}_2^*) \end{aligned} \quad (12)$$

Substituting on the left-hand side of equation 11:

$$\begin{aligned} f_1(f_2(x, \text{pa}_2, \text{pa}_2^*), \text{pa}_1, \text{pa}_1^*) &= f_1(g(\epsilon, \text{pa}_1, \text{pa}_2^*), \text{pa}_1, \text{pa}_1^*) \\ &= g(\epsilon, \text{pa}_1^*, \text{pa}_2^*) \end{aligned} \quad (13)$$

And on the right-hand side:

$$\begin{aligned} f_2(f_1(x, \text{pa}_1, \text{pa}_1^*), \text{pa}_2, \text{pa}_2^*) &= f_2(g(\epsilon, \text{pa}_1^*, \text{pa}_2), \text{pa}_2, \text{pa}_2^*) \\ &= g(\epsilon, \text{pa}_1^*, \text{pa}_2^*) \end{aligned} \quad (14)$$

□

## D  COLOUR MNIST

### D.1  DATASET DETAILS

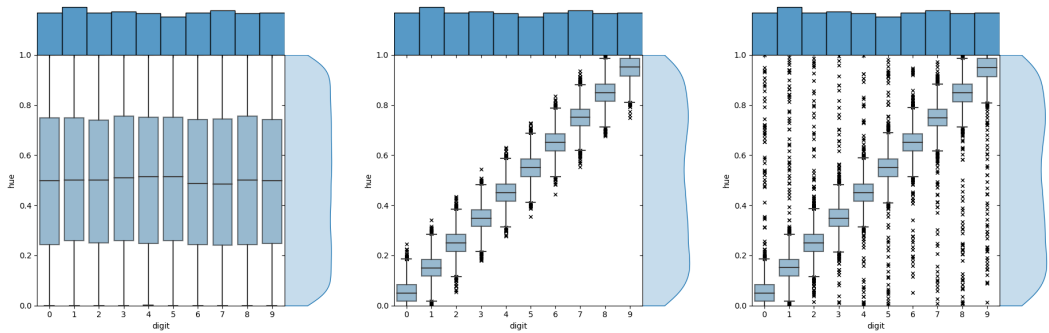

(a) Unconfounded joint distribution.  (b) Confounded joint distribution without full support.  (c) Confounded joint distribution with full support.

Figure 4: Colour MNIST joint distribution of digit and hue for different SCMs.

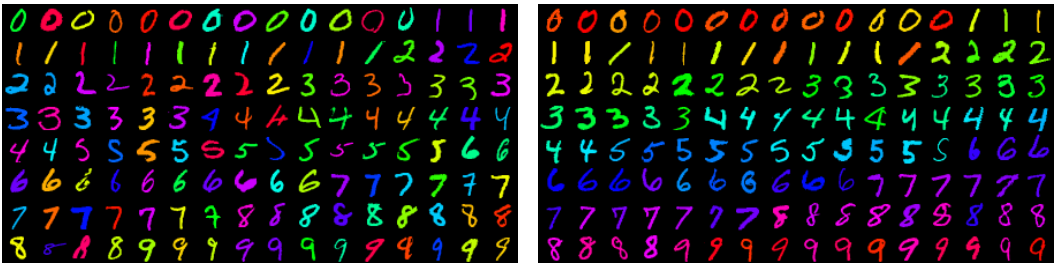

(a) Samples from the unconfounded joint distribution.  (b) Samples from the confounded joint distribution.

Figure 5: Colour MNIST samples.

### D.2  EXPERIMENT DETAILS

We describe the architectures of the different components of the model using pseudo-code snippets. These snippets are close to the JAX framework but should enable reproducing the same architectures in all popular deep learning frameworks.

#### D.2.1  PSEUDO-ORACLES.

**Architecture.**   The architecture of the pseudo-oracles is:

```
pseudo_oracle = serial(
    Conv(out_chan=64, filter_shape=(4, 4), strides=(2, 2)),
    LeakyRelu,
    Conv(out_chan=64, filter_shape=(4, 4), strides=(2, 2)),
    LeakyRelu,
    Flatten,
    Dense(out_dim=128),
    LeakyRelu,
    Dense(out_dim=num_classes if classification else 1)
)
y_hat = pseudo_oracle(image)
```

**Training details.**   We trained the pseudo-oracles for 2000 steps with a batch size of 1024 using the AdamW (Loshchilov & Hutter, 2019) optimiser with a learning rate of 0.0005, $\beta_1 = 0.9$, $\beta_2 = 0.999$ and weight_decay $= 0.0001$. For data augmentation we used random translations where the amount of horizontal and vertical translation is sampled uniformly from 0 to 10% of the width/height of the image.

### D.2.2 VAE

The following image encoder and decoder are used in both the GAN and VAE:

**Generic image encoder and decoder.**

```
image_encoder = serial(
    Conv(out_chan=64, filter_shape=(4, 4), strides=(2, 2)),
    LeakyRelu,
    Conv(out_chan=64, filter_shape=(4, 4), strides=(2, 2)),
    LeakyRelu,
    Flatten,
    Dense(out_dim=128),
    LeakyRelu)
)

image_decoder = serial(
    Dense(7 * 7 * 64),
    LeakyRelu,
    Reshape((-1, 7, 7, 64)),
    Resize((-1, 14, 14, 64), method='linear'),
    Conv(out_chan=64, filter_shape=(4, 4), strides=(1, 1), padding='SAME'),
    LeakyRelu,
    Resize((-1, 28, 28, 64), method='linear'),
    Conv(out_chan=64, filter_shape=(4, 4), strides=(1, 1), padding='SAME'),
    LeakyRelu,
    Conv(out_chan=3, filter_shape=(3, 3), strides=(1, 1), padding='SAME'))
```

**Architecture.** The VAE architecture is as follows:

```
#encoder
encoded_image = image_encoder(image)
tmp = Dense(out_dim=128)(concat(encoded_image, parents))
mu = Dense(out_dim=16)(tmp)
sigma = softplus(Dense(out_dim=16)(tmp))
#decoder
z = sample_from_standard_normal(mu, sigma)
counterfactual = image_decoder(concat(z, counterfactual_parents))
```

**Training details.** We trained the VAE for 20000 steps with a batch size of 512 using the AdamW (Loshchilov & Hutter, 2019) optimiser with a learning rate of 0.001, $\beta_1 = 0.9$, $\beta_2 = 0.999$ and weight_decay $= 0.0001$. The data-augmentation is the same as in Section D.2.1.

### D.2.3 GAN

**Architecture.** The architecture of the GAN generator and discriminator are as follows:

```
#generator
gan_decoder = serial(
            Dense(out_dim=128),
            LeakyRelu,
            image_decoder,
            Tanh
        )
encoded_image = image_encoder(image)
```

```
tmp = concat(encoded_image, parents, counterfactual_parents)
counterfactual = gan_decoder(tmp)

# critic
critic = serial(
        Conv(out_chan=64, filter_shape=(4, 4), strides=(2, 2), padding='SAME'),
        LeakyRelu,
        Conv(out_chan=64, filter_shape=(4, 4), strides=(2, 2), padding='SAME'),
        LeakyRelu,
        Flatten,
        Dense(out_dim=128),
        LeakyRelu,
        Dense(out_dim=128),
        LeakyRelu
        Dense(1)
    )
)
critic_input = concat(image, broadcast_to_shape(parents, image.shape))
logits = critic(critic_input)
```

**Training details.** We trained the GAN for 20000 steps with a batch size of 512 using the AdamW (Loshchilov & Hutter, 2019) optimiser with a learning rate of 0.0001, $\beta_1 = 0$, $\beta_2 = 0.9$ and weight_decay $= 0.0001$. The learning rate is multiplied by 0.5 at 12000 steps and again multiplied by 0.2 at 16000 steps. The data-augmentation is the same as in Section D.2.1.

### D.3 EXTRA RESULTS

Table 2: Effectiveness results on colour MNIST when using pseudo oracles trained from biased data without using a simulated intervention. We measure effectiveness using digit accuracy and hue absolute error in percentage points since $hue \in [0, 1]$. The average model ranking change compared to Table 1 (oracles trained from unbiased data) was 0.8(3) for the dataset without full support and 0.75 for the dataset with full support.

| dataset | intervention | model | digit intervention effectiveness | | hue intervention effectiveness | | avg. rank | original rank | abs. diff. |
|---|---|---|---|---|---|---|---|---|---|
| | | | $\text{acc}_{\text{digit}}(\%) \uparrow$ | $\text{ae}_{\text{hue}}(\%) \downarrow$ | $\text{acc}_{\text{digit}}(\%) \uparrow$ | $\text{ae}_{\text{hue}}(\%) \downarrow$ | | | |
| confounded w/o full support | no | Bernoulli VAE $\beta$=1 | 53.77 (9.88) | 11.43 (4.56) | 53.68 (6.08) | 11.80 (5.85) | 3.5 | 3.5 | 0 |
| | | Bernoulli VAE $\beta$=2 | 48.34 (2.27) | 10.42 (1.64) | 49.32 (2.01) | 10.15 (1.62) | 4 | 2.5 | 1.5 |
| | | Normal VAE $\beta$=5 | 57.65 (8.41) | 13.92 (3.38) | 60.69 (6.96) | 13.63 (3.51) | 3.5 | 5.5 | 2 |
| | yes | Bernoulli VAE $\beta$=1 | 49.83 (2.52) | 9.15 (0.87) | 50.92 (2.07) | 9.51 (1.35) | 3 | 2.25 | 0.75 |
| | | Bernoulli VAE $\beta$=2 | 46.48 (1.22) | 8.51 (0.87) | 47.30 (1.34) | 8.53 (0.87) | 3.5 | 3.25 | 0.25 |
| | | Normal VAE $\beta$=5 | 43.86 (1.28) | 8.28 (1.38) | 47.19 (1.10) | 7.97 (1.34) | 3.5 | 4 | 0.5 |
| confounded w/ full support | no | Bernoulli VAE $\beta$=1 | 87.71 (0.92) | 10.90 (6.53) | 90.15 (0.48) | 14.02 (10.59) | 7 | 6.5 | 0.5 |
| | | Bernoulli VAE $\beta$=2 | 90.05 (0.44) | 5.94 (0.20) | 90.64 (0.52) | 5.92 (0.23) | 5.25 | 3.25 | 2 |
| | | Normal VAE $\beta$=5 | 82.98 (0.91) | 6.19 (0.73) | 89.23 (0.19) | 5.77 (0.65) | 6.5 | 7 | 0.5 |
| | yes* | GAN | 87.03 (0.62) | 4.71 (0.16) | 92.51 (0.36) | 5.28 (0.34) | 3 | 4.25 | 1.25 |
| | yes | Bernoulli VAE $\beta$=1 | 92.07 (0.25) | 5.64 (0.19) | 92.19 (0.35) | 6.15 (0.59) | 3.75 | 4 | 0.25 |
| | | Bernoulli VAE $\beta$=2 | 93.98 (0.26) | 5.72 (0.25) | 93.76 (0.15) | 5.82 (0.23) | 2.5 | 2 | 0.5 |
| | | Normal VAE $\beta$=5 | 88.68 (0.33) | 5.92 (0.46) | 91.85 (0.12) | 5.87 (0.47) | 4.5 | 4.75 | 0.25 |
| | | GAN | 87.40 (0.14) | 4.78 (0.18) | 90.90 (0.21) | 4.90 (0.16) | 3.5 | 4.25 | 0.75 |

Table 3: Effectiveness results on colour MNIST when using pseudo oracles trained from biased data using a simulated intervention. We measure effectiveness using digit accuracy and hue absolute error in percentage points since $hue \in [0, 1]$. The average model ranking change compared to Table 1 (oracles trained from unbiased data) was 0.8(3) for the dataset without full support and 0.625 for the dataset with full support.

| dataset | intervention | model | digit intervention effectiveness | | hue intervention effectiveness | | avg. rank | original rank | abs. diff. |
| --- | --- | --- | --- | --- | --- | --- | --- | --- | --- |
| | | | $acc_{digit}(\%) \uparrow$ | $ae_{hue}(\%) \downarrow$ | $acc_{digit}(\%) \uparrow$ | $ae_{hue}(\%) \downarrow$ | | | |
| confounded w/o full support | no | Bernoulli VAE $\beta$=1 | 57.09 (10.01) | 11.17 (4.88) | 57.14 (6.63) | 11.42 (6.06) | 3.5 | 3.5 | 0 |
| | | Bernoulli VAE $\beta$=2 | 53.43 (2.44) | 10.08 (1.57) | 54.03 (2.57) | 9.73 (1.55) | 4 | 2.5 | 1.5 |
| | | Normal VAE $\beta$=5 | 62.93 (8.26) | 13.81 (3.27) | 67.23 (7.21) | 13.55 (3.38) | 3.5 | 5.5 | 2 |
| | yes | Bernoulli VAE $\beta$=1 | 53.72 (2.61) | 8.57 (1.00) | 55.36 (2.37) | 8.88 (1.45) | 3 | 2.25 | 0.75 |
| | | Bernoulli VAE $\beta$=2 | 50.96 (0.76) | 7.93 (0.96) | 51.79 (0.72) | 7.89 (0.90) | 3.5 | 3.25 | 0.25 |
| | | Normal VAE $\beta$=5 | 47.89 (1.75) | 7.83 (1.58) | 51.23 (2.01) | 7.42 (1.59) | 3.5 | 4 | 0.5 |
| confounded w/ full support | no | Bernoulli VAE $\beta$=1 | 93.25 (1.48) | 8.93 (7.08) | 96.57 (0.42) | 12.36 (11.61) | 7.5 | 6.5 | 1 |
| | | Bernoulli VAE $\beta$=2 | 96.90 (0.40) | 3.17 (0.11) | 96.94 (0.20) | 2.98 (0.14) | 4.25 | 3.25 | 1 |
| | | Normal VAE $\beta$=5 | 90.37 (0.45) | 4.16 (1.07) | 95.92 (0.25) | 3.85 (1.11) | 7.5 | 7 | 0.5 |
| | yes* | GAN | 93.30 (0.69) | 2.36 (0.19) | 97.47 (0.23) | 2.52 (0.09) | 3.25 | 4.25 | 1 |
| | yes | Bernoulli VAE $\beta$=1 | 97.36 (0.18) | 2.80 (0.14) | 97.70 (0.25) | 3.67 (0.83) | 3.5 | 4 | 0.5 |
| | | Bernoulli VAE $\beta$=2 | 98.74 (0.11) | 2.68 (0.17) | 98.84 (0.16) | 2.61 (0.16) | 2 | 2 | 0 |
| | | Normal VAE $\beta$=5 | 94.68 (0.39) | 3.26 (0.24) | 97.50 (0.21) | 3.20 (0.25) | 4.5 | 4.75 | 0.25 |
| | | GAN | 93.90 (0.18) | 2.38 (0.16) | 96.58 (0.15) | 2.44 (0.17) | 3.5 | 4.25 | 0.75 |

Table 4: Effectiveness on colour MNIST when using linear/logistic regression as pseudo oracles trained from unbiased data. We measure effectiveness using digit accuracy and hue absolute error in percentage points since $hue \in [0, 1]$. The average model ranking change compared to Table 1 (oracles trained from unbiased data) was 0.91(6) for the dataset without full support and 0.6875 for the dataset with full support.

| dataset | intervention | model | digit intervention effectiveness | | hue intervention effectiveness | | avg. rank | original rank | abs. diff. |
| --- | --- | --- | --- | --- | --- | --- | --- | --- | --- |
| | | | $acc_{digit}(\%) \uparrow$ | $ae_{hue}(\%) \downarrow$ | $acc_{digit}(\%) \uparrow$ | $ae_{hue}(\%) \downarrow$ | | | |
| confounded w/o full support | no | Bernoulli VAE $\beta$=1 | 80.83 (1.91) | 17.77 (4.12) | 81.79 (1.42) | 17.28 (4.14) | 4.75 | 3.5 | 1.25 |
| | | Bernoulli VAE $\beta$=2 | 86.18 (0.63) | 16.17 (1.32) | 86.03 (0.52) | 15.91 (1.27) | 2.5 | 2.5 | 0 |
| | | Normal VAE $\beta$=5 | 78.67 (4.32) | 18.97 (1.56) | 82.43 (3.80) | 18.37 (1.22) | 5.25 | 5.5 | 0.25 |
| | yes | Bernoulli VAE $\beta$=1 | 82.23 (0.81) | 15.22 (0.63) | 82.31 (1.01) | 14.24 (0.74) | 2.75 | 2.25 | 0.5 |
| | | Bernoulli VAE $\beta$=2 | 80.52 (2.11) | 15.93 (1.34) | 79.37 (2.49) | 15.74 (1.41) | 4.25 | 3.25 | 1 |
| | | Normal VAE $\beta$=5 | 82.50 (1.61) | 14.17 (1.01) | 85.20 (1.13) | 13.82 (1.08) | 1.5 | 4 | 2.5 |
| confounded w/ full support | no | Bernoulli VAE $\beta$=1 | 82.26 (2.97) | 18.05 (4.15) | 84.23 (1.16) | 18.73 (5.29) | 7.25 | 6.5 | 0.75 |
| | | Bernoulli VAE $\beta$=2 | 87.55 (0.47) | 14.75 (0.19) | 87.00 (0.28) | 14.80 (0.20) | 3.5 | 3.25 | 0.25 |
| | | Normal VAE $\beta$=5 | 81.78 (0.93) | 15.14 (0.39) | 84.90 (0.59) | 14.36 (0.44) | 5.25 | 7 | 1.75 |
| | yes* | GAN | 80.01 (1.28) | 14.71 (0.08) | 87.09 (0.84) | 14.87 (0.17) | 5 | 4.25 | 0.75 |
| | yes | Bernoulli VAE $\beta$=1 | 87.17 (0.64) | 14.88 (0.11) | 86.09 (1.73) | 14.37 (0.64) | 3.75 | 4 | 0.25 |
| | | Bernoulli VAE $\beta$=2 | 89.87 (0.51) | 14.70 (0.20) | 89.32 (0.34) | 14.56 (0.16) | 1.5 | 2 | 0.5 |
| | | Normal VAE $\beta$=5 | 85.43 (0.59) | 15.00 (0.17) | 87.23 (0.53) | 14.84 (0.25) | 4.5 | 4.75 | 0.25 |
| | | GAN | 80.39 (0.56) | 14.77 (0.15) | 85.51 (0.51) | 14.65 (0.12) | 5.25 | 4.25 | 1 |

# E  3D SHAPES

## E.1  EXPERIMENT DETAILS

For the 3D Shapes dataset we used the same architectures and training regimes as in Section D.2 except we did not use data-augmentation and modified the image encoder and decoder as follows:

```
image_encoder = serial(
    Conv(out_chan=64, filter_shape=(4, 4), strides=(2, 2)),
    LeakyRelu,
    Conv(out_chan=64, filter_shape=(4, 4), strides=(2, 2)),
    LeakyRelu,
    Conv(out_chan=64, filter_shape=(4, 4), strides=(2, 2)),
    LeakyRelu,
    Flatten,
    Dense(out_dim=128),
    LeakyRelu)
)

image_decoder = serial(
    Dense(8 * 8 * 64),
    LeakyRelu,
    Reshape((-1, 8, 8, 64)),
    Resize((-1, 16, 16, 64), method='linear'),
    Conv(out_chan=64, filter_shape=(4, 4), strides=(1, 1), padding='SAME'),
    LeakyRelu,
    Resize((-1, 32, 32, 64), method='linear'),
    Conv(out_chan=64, filter_shape=(4, 4), strides=(1, 1), padding='SAME'),
    LeakyRelu,
    Resize((-1, 64, 64, 64), method='linear'),
    Conv(out_chan=64, filter_shape=(4, 4), strides=(1, 1), padding='SAME'),
    LeakyRelu,
    Conv(out_chan=3, filter_shape=(3, 3), strides=(1, 1), padding='SAME'))
```

## E.2  EXTRA RESULTS

Table 5: Results of quality tests on 3D shapes: we measure composition after the null intervention and reversibility after one intervention cycle. Effectiveness is measured using accuracy.

| intervention | model | effectiveness (%) | | | | | | reversibility | |
|---|---|---|---|---|---|---|---|---|---|
| | | $\text{acc}_{\text{floorhue}}$ | $\text{acc}_{\text{objecthue}}$ | $\text{acc}_{\text{objectorientation}}$ | $\text{acc}_{\text{objectscale}}$ | $\text{acc}_{\text{objectshape}}$ | $\text{acc}_{\text{wallhue}}$ | $l_1^{(1)}$ | $l_1^{(10)}$ |
| floor hue | VAE | 100.00 | 100.00 | 100.00 | 100.00 | 100.00 | 100.00 | 1.62 (0.07) | 1.62 (0.07) |
| | GAN | 100.00 | 100.00 | 99.90 (0.16) | 99.37 (0.13) | 99.30 (0.16) | 100.00 | 5.23 (0.15) | 5.25 (0.15) |
| object hue | VAE | 100.00 | 100.00 | 100.00 | 100.00 | 100.00 | 100.00 | 1.62 (0.07) | 1.62 (0.07) |
| | GAN | 100.00 | 99.99 (0.01) | 99.91 (0.15) | 99.29 (0.16) | 99.16 (0.25) | 100.00 | 5.33 (0.16) | 5.35 (0.16) |
| object orientation | VAE | 100.00 | 100.00 | 100.00 | 100.00 | 100.00 | 100.00 | 1.62 (0.07) | 1.62 (0.07) |
| | GAN | 100.00 | 100.00 | 99.88 (0.16) | 99.07 (0.15) | 99.00 (0.22) | 100.00 | 5.32 (0.14) | 5.35 (0.14) |
| object scale | VAE | 100.00 | 100.00 | 100.00 | 100.00 | 100.00 | 100.00 | 1.62 (0.07) | 1.62 (0.07) |
| | GAN | 100.00 | 100.00 | 99.91 (0.15) | 98.83 (0.27) | 99.19 (0.27) | 100.00 | 5.30 (0.15) | 5.31 (0.15) |
| object shape | VAE | 100.00 | 100.00 | 100.00 | 100.00 | 100.00 | 100.00 | 1.62 (0.07) | 1.62 (0.07) |
| | GAN | 100.00 | 100.00 | 99.92 (0.12) | 98.76 (0.27) | 99.10 (0.16) | 100.00 | 5.37 (0.15) | 5.39 (0.15) |
| wall hue | VAE | 100.00 | 100.00 | 100.00 | 100.00 | 100.00 | 100.00 | 1.62 (0.07) | 1.62 (0.07) |
| | GAN | 100.00 | 100.00 | 99.86 (0.20) | 99.30 (0.19) | 99.32 (0.19) | 100.00 | 5.27 (0.16) | 5.29 (0.16) |

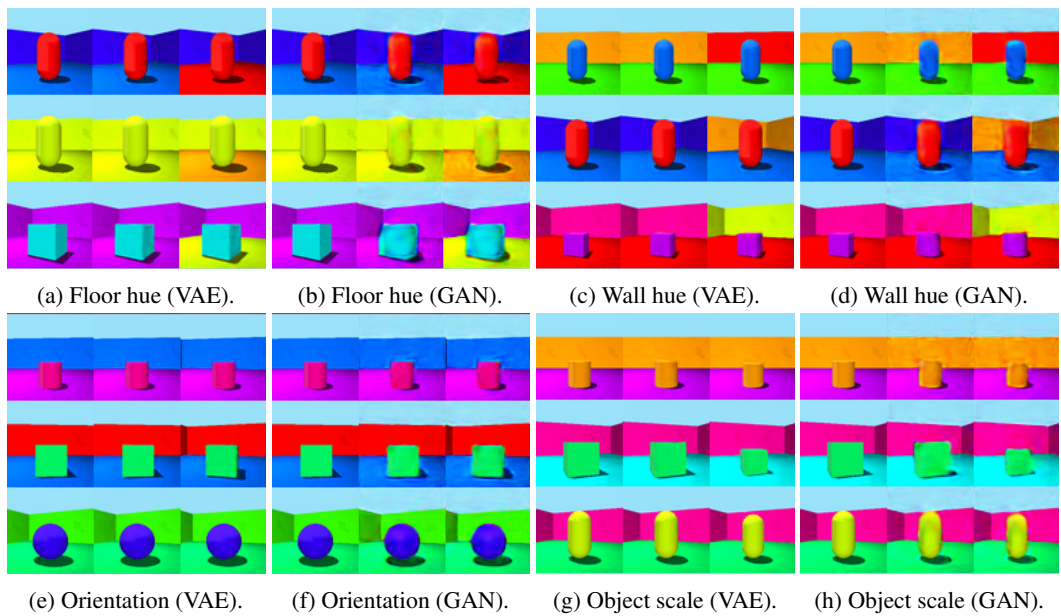

(a) Floor hue (VAE).  (b) Floor hue (GAN).  (c) Wall hue (VAE).  (d) Wall hue (GAN).

(e) Orientation (VAE).  (f) Orientation (GAN).  (g) Object scale (VAE).  (h) Object scale (GAN).

Figure 6: Effectiveness/Composition on 3D shapes. Each sub-figure has one sample per row: the first column is the input, the second column is the result of the null intervention, and the last column is the result of a partial intervention on a specific parent.

# F CELEBA-HQ

## F.1 DATASET DETAILS

For our CelebA-HQ (Karras et al., 2018) experiments, we randomly split the 30,000 examples into 70% for training, 15% for validation and 15% for testing. We selected the 'smiling' and 'eyeglasses' binary parent attributes for conditioning our generative model. After careful consideration, we opted for these attributes for three main reasons: (i) they are gender neutral which mitigates bias and allows us to use the entire dataset rather than a smaller subset; (ii) they are more objective than other attributes like 'attractive' and 'young'; (iii) we can train reasonably accurate pseudo-oracles for our counterfactual evaluation on these attributes, which is not the case for some others due to label noise/ambiguity and significant class imbalance.

## F.2 EXPERIMENT DETAILS

### F.2.1 GENERATIVE MODEL

**Architecture.** For the experiments on the CelebA-HQ dataset, we follow the general very deep VAE (VDVAE) setup proposed by (Child, 2021), and introduce some modifications to accommodate both parent conditioning and our compute constraints. The architecture is based on the ResNet VAE proposed by (Kingma et al., 2016) but is much deeper and uses bottleneck residual blocks. The VDVAE has several stochastic layers of latent variables which are conditionally dependent upon each other, and are organised into $L$ groups $\mathbf{z} = \{\mathbf{z}_0, \mathbf{z}_1, \ldots, \mathbf{z}_L\}$. These latents are typically output as feature maps of varying resolutions, whereby $\mathbf{z}_0$ consists of fewer latents at low resolution up to many latents $\mathbf{z}_L$ at high resolution. The conditioning structure is organised following the ladder structure proposed by (Sønderby et al., 2016), where both the prior $p_\theta(\mathbf{z})$ and approximate posterior $q_\phi(\mathbf{z}|\mathbf{x})$ generate latent latent variables in the same top-down order. As in previous work, the prior and posterior are diagonal Gaussian distributions and the model is trained end-to-end by optimizing the usual variational bound on the log-likelihood (ELBO) (Kingma & Welling, 2013; Maaløe et al., 2019). We found that a naive application of a VDVAE to produce high resolution counterfactuals leads to ignored counterfactual parent conditioning. Although the setup we describe next worked well enough in our experiments, understanding and overcoming this issue in the general case likely warrants further investigation.

For our experiments, we used a VDVAE with stochastic latent variables spanning 6 resolution scales up to the $64 \times 64$ input resolution: $\{1^2, 4^2, 8^2, 16^2, 32^2, 64^2\}$, where each latent variable has 16 channels. We used the following number of residual blocks per resolution scale: $\{4, 4, 8, 12, 12, 4\}$, resulting in a total of 42 stochastic latent variable layers and 19M trainable parameters. Additionally, we modified the original architecture from a fixed channel width (e.g. 384) across all resolutions to the following custom channel widths per resolution: $\{32, 64, 128, 256, 512, 1024\}$. We found that reducing the number of channels at higher resolutions and increasing them for lower resolutions performed well enough in our experiments, whilst reducing both memory and runtime requirements significantly.

**Conditional VDVAE.** For our counterfactual generation purposes, we augmented the original prior and posterior top-down conditioning structure to include $\mathbf{x}$'s parents **pa** as follows:

$$p_\theta(\mathbf{z}|\mathbf{pa}) = p_\theta(\mathbf{z}_0) \prod_{i=1}^L p_\theta(\mathbf{z}_i|\mathbf{z}_{i-1}, \mathbf{pa}), \quad q_\phi(\mathbf{z}|\mathbf{x}, \mathbf{pa}) = q_\phi(\mathbf{z}_0|\mathbf{x}) \prod_{i=1}^L q_\phi(\mathbf{z}_i|\mathbf{z}_{i-1}, \mathbf{x}, \mathbf{pa}). \quad (15)$$

We call this the Conditional VDVAE. In practical terms, we simply expand and concatenate the parent conditioning attributes to the latent variables at each stochastic layer, then merge them into the downstream via a $1 \times 1$ convolution. Although there are many other conditioning structures one could consider, this one proved good enough for our experiments. In order to produce counterfactuals, we perform abduction by passing images and their parent attributes through the encoder and retrieving the posterior latent variables at each stochastic layer. We then fix these latents and propagate them through the decoder along with the counterfactual parent conditioning. We also found it beneficial to replace the original diagonal discretized logistic mixture likelihood (Salimans et al., 2017) used in VDVAEs with a diagonal discretized Gaussian likelihood (Ho et al., 2020), as it produced visually sharper counterfactuals in our experiments. Importantly, our model was trained with a $\beta$ penalty (Higgins et al., 2017) of 5 which discouraged it from focusing mostly on maximising the likelihood term in the ELBO and ignoring counterfactual conditioning at inference time. We found that this introduces a trade-off between reconstruction quality and obeying counterfactual conditioning.

**Training Details.** We trained our VDVAE for 1.7M steps with a batch size of 32 using the AdamW (Loshchilov & Hutter, 2019) optimiser with an initial learning rate of 1e-3, $\beta_1 = 0.9$, $\beta_2 = 0.9$ and a weight decay of $0.01$. The learning rate was linearly warmed-up from 0 to 1e-3 over the first 100 steps then reduced to 1.5e-4 at 175K steps and again to 1.5e-5 at 900K steps. We set gradient clipping to 220 and gradient L2 norm skipping threshold to 380, and observed no significant instabilities during training. The final artefact is an exponential moving average of the model parameters with a rate of $0.999$ which we use at inference time. For data-augmentation, since the initial dataset was $128 \times 128$ resolution, during training we applied random cropping to $120 \times 120$ and resized to $64 \times 64$ before applying random horizontal flipping with probability $0.5$ and finally scaling pixel intensities to $[-1, 1]$. At test time we simply resize to $64 \times 64$.

### F.2.2 PSEUDO-ORACLES

**Architecture.** In order to evaluate the effectiveness of our CelebA-HQ counterfactuals, we trained the following simple classifier for each of the 'smiling' and 'eyeglasses' binary attributes:

```python
class ConvNet(nn.Module):
    def __init__(self):
        super().__init__()
        self.cnn = nn.Sequential(
            nn.Conv2d(3, 16, 3, 1, 1),
            nn.BatchNorm2d(16),
            nn.ReLU(),
            nn.Conv2d(16, 32, 3, 2, 1),
            nn.BatchNorm2d(32),
            nn.ReLU(),
            nn.Conv2d(32, 32, 3, 1, 1),
            nn.BatchNorm2d(32),
            nn.ReLU(),
            nn.Conv2d(32, 64, 3, 2, 1),
            nn.BatchNorm2d(64),
            nn.ReLU(),
            nn.Conv2d(64, 64, 3, 1, 1),
            nn.BatchNorm2d(64),
            nn.ReLU(),
            nn.Conv2d(64, 128, 3, 2, 1),
            nn.BatchNorm2d(128),
            nn.ReLU(),
            nn.AdaptiveAvgPool2d(1),
        )
        self.fc = nn.Sequential(
            nn.Linear(128, 128),
            nn.BatchNorm1d(128),
            nn.ReLU(),
            nn.Linear(128, 1)
        )

    def forward(self, x):
        return self.fc(self.cnn(x).squeeze())
```

**Training Details.** Our attribute classifiers were trained for 300 epochs with a batch size of 32 using the AdamW (Loshchilov & Hutter, 2019) optimizer with an initial learning rate of 1e-3, $\beta_1 = 0.9$, $\beta_2 = 0.999$ and weight decay of $0.01$. The learning rate followed a plateau schedule on the validation set F1-score, where it was halved after 50 epochs of no improvement. The final artefact is an exponential moving average of the model parameters with a rate of 0.99 which we use at inference time. To find the best binary classification threshold between 0 and 1, we evaluated the validation set at threshold intervals of 0.001 and took the best performing threshold in terms of F1-score. We ran training from 3 random seeds and the final test set F1-score for the 'smiling' attribute is $0.9342 \pm 0.0018$, and for the 'eyeglasses' attribute it is $0.9758 \pm 0.0037$. For data augmentation, we followed the VDVAE setup but also included the torchvision.transforms.AutoAugment() augmentation module from PyTorch (Paszke et al., 2019) during training to strengthen regularisation. Since the CelebA-HQ binary attributes can be highly imbalanced, we also used a weighted random data sampler (with replacement) during training, which ensures minibatches have balanced representation of each class at each step.

### F.3 EXTRA RESULTS

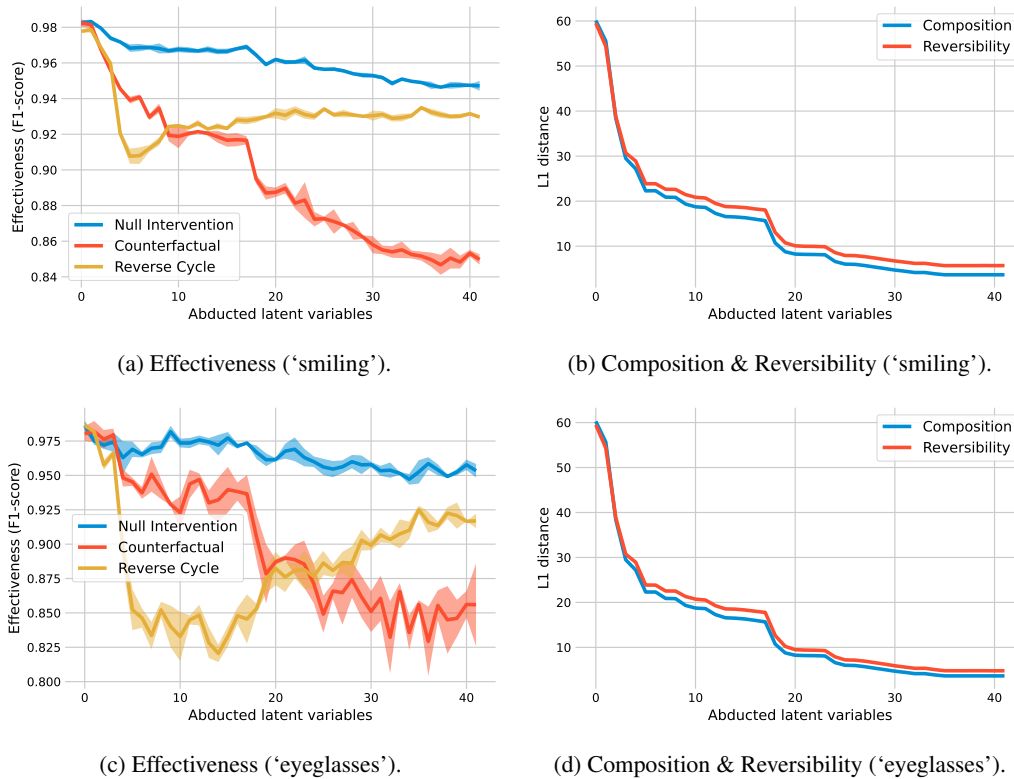

(a) Effectiveness ('smiling').

(b) Composition & Reversibility ('smiling').

(c) Effectiveness ('eyeglasses').

(d) Composition & Reversibility ('eyeglasses').

Figure 7: Measuring CelebA-HQ test set effectiveness, composition and reversibility as a function of the number of latent variables abducted. Notice how counterfactual effectiveness drops rapidly as more (posterior) latent variables are abducted, unlike the null intervention and reverse cycle case.

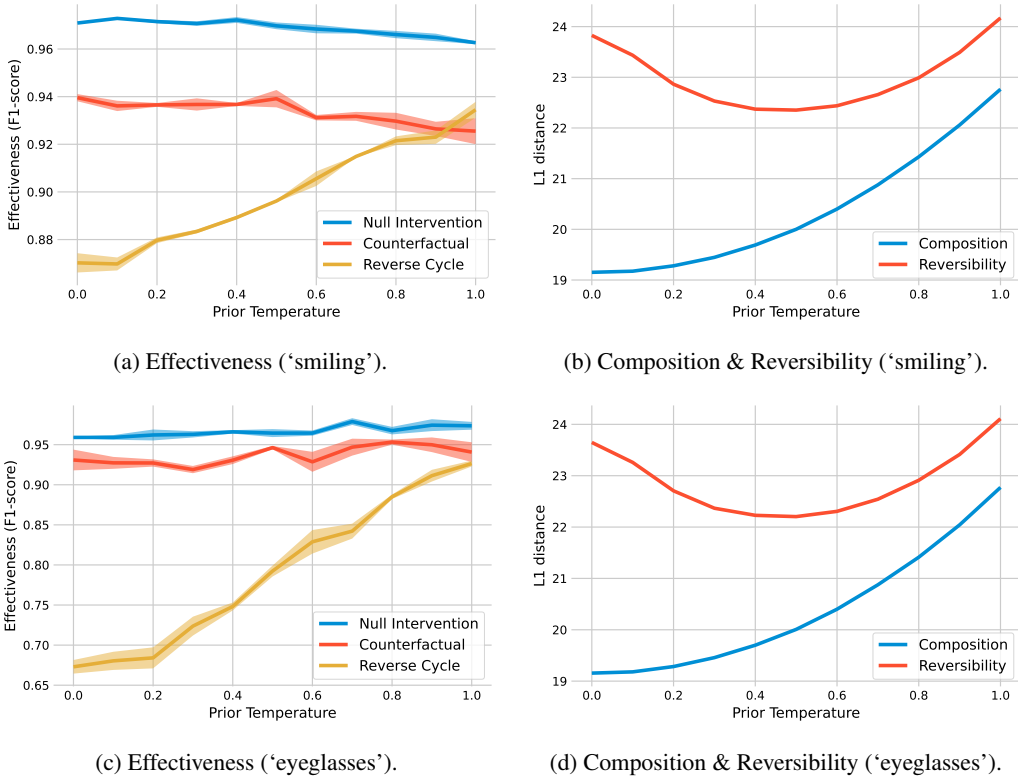

(a) Effectiveness ('smiling').

(b) Composition & Reversibility ('smiling').

(c) Effectiveness ('eyeglasses').

(d) Composition & Reversibility ('eyeglasses').

Figure 8: Measuring CelebA-HQ test set effectiveness, composition and reversibility as a function of the temperature used when sampling from the prior. In this example, the first 8 (posterior) latent variables were abducted and the remaining 34 were randomly sampled from the prior. Notice how low prior temperatures affect the effectiveness of cycled back images the most (Figure (a)).

Table 6: Soundness metrics on the CelebA-HQ test set over 3 random seeds, with different subsets of abducted latent variables from our conditional VDVAE model. Composition is measured via the null intervention and reversibility after one intervention cycle. Effectiveness of test set counterfactuals is measured using F1-score given by our 'smiling'/'eyeglasses' attribute classifiers.

| | smiling intervention | | | eyeglasses intervention | | |
|---|---|---|---|---|---|---|
| latents abducted | composition $l_1^{(1)} \downarrow$ | reversibility $l_1^{(1)} \downarrow$ | effectiveness F1-score $\uparrow$ | composition $l_1^{(1)} \downarrow$ | reversibility $l_1^{(1)} \downarrow$ | effectiveness F1-score $\uparrow$ |
| 1 | 60.100 (0.127) | 59.521 (0.164) | 0.984 (0.0005) | 60.273 (0.144) | 59.263 (0.013) | 0.979 (0.002) |
| 2 | 55.521 (0.068) | 54.360 (0.092) | 0.982 (0.0008) | 55.443 (0.036) | 54.314 (0.055) | 0.982 (0.005) |
| 4 | 29.489 (0.018) | 30.705 (0.030) | 0.957 (0.001) | 29.491 (0.048) | 30.758 (0.025) | 0.976 (0.005) |
| 8 | 20.890 (0.018) | 22.604 (0.025) | 0.933 (0.002) | 20.896 (0.011) | 22.526 (0.005) | 0.932 (0.006) |
| 16 | 15.997 (0.003) | 18.269 (0.021) | 0.921 (0.003) | 16.312 (0.009) | 18.323 (0.009) | 0.932 (0.019) |
| 24 | 6.606 (0.001) | 9.880 (0.017) | 0.886 (0.006) | 8.092 (0.002) | 9.290 (0.006) | 0.872 (0.003) |
| 32 | 4.160 (0.002) | 6.189 (0.015) | 0.858 (0.001) | 4.451 (0.0006) | 5.606 (0.010) | 0.865 (0.023) |
| 42 (all) | 3.657 (0.0006) | 5.701 (0.017) | 0.848 (0.0006) | 3.656 (0.001) | 4.808 (0.004) | 0.854 (0.006) |

Table 7: Ablation of $\beta$ penalty training on composition, reversibility and effectiveness. Soundness metrics were calculated on the CelebA-HQ test set using the 'smiling' attribute, over 3 random seeds. We used our conditional VDVAE with a diagonal discretized logistic mixture likelihood. Notice how no penalty ($\beta=1$) yields the best likelihood, composition and reversibility at the cost of low effectiveness (i.e. ignored counterfactual conditioning under full abduction).

| $\beta$ penalty | latents abducted | NLL bits per dim ↓ | composition $l_1^{(1)}$ ↓ | reversibility $l_1^{(1)}$ ↓ | effectiveness F1-score ↑ |
|---|---|---|---|---|---|
| 1 | 42 | ≤ 2.874 | 1.794 (0.0002) | 2.482 (0.002) | 0.536 (0.014) |
| 3 | 42 | ≤ 3.556 | 4.335 (0.001) | 5.333 (0.002) | 0.720 (0.002) |
| 5 | 42 | ≤ 3.931 | 6.362 (0.002) | 7.775 (0.013) | 0.877 (0.002) |
| 7 | 42 | ≤ 4.189 | 10.020 (0.008) | 8.040 (0.0006) | 0.910 (0.003) |
| 10 | 42 | ≤ 4.446 | 12.209 (0.003) | 10.050 (0.002) | 0.952 (0.001) |

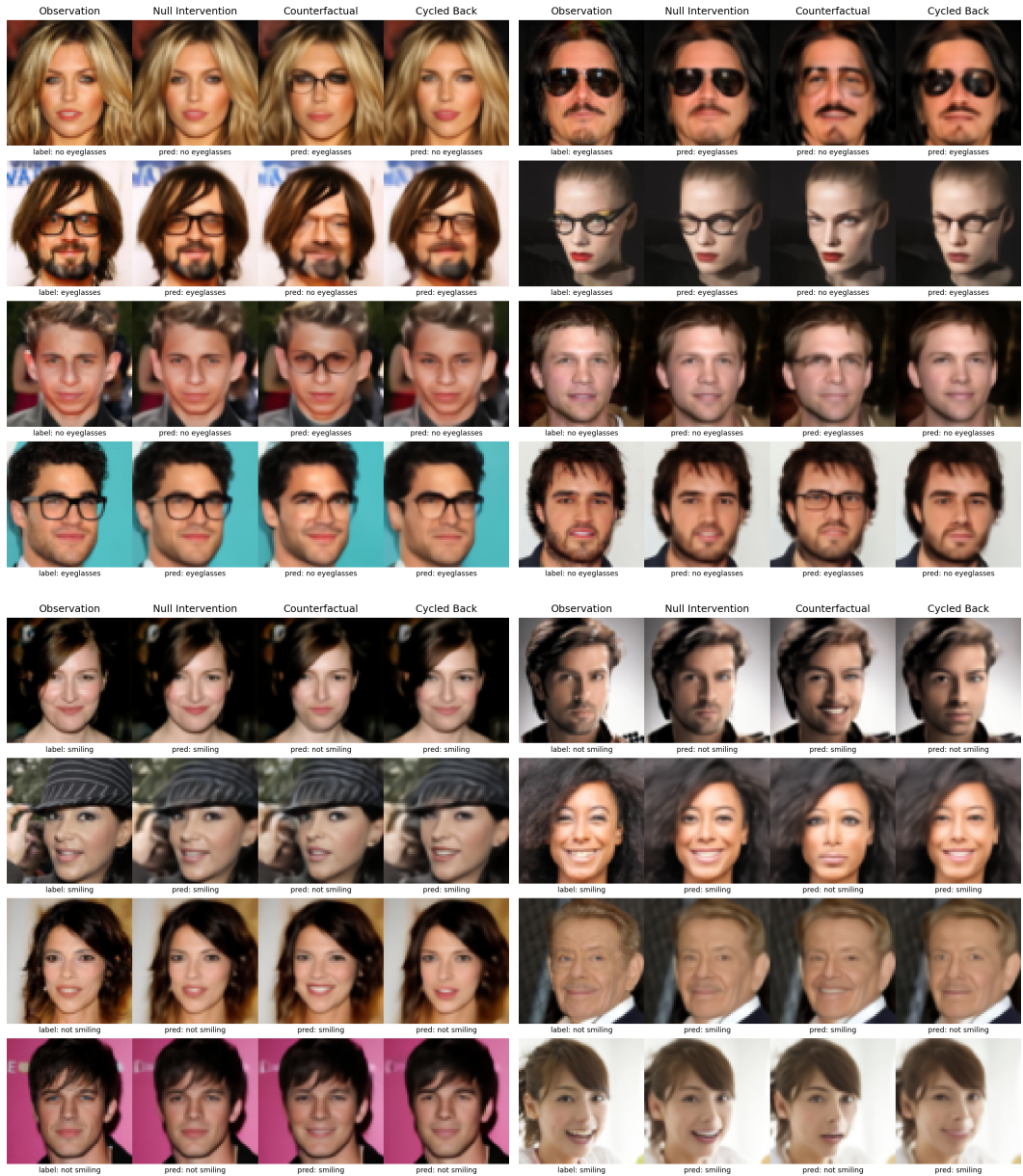

Figure 9: CelebA-HQ test set counterfactuals of 'eyeglasses' (Top) and 'smiling' (Bottom) attributes. 'Null Intervention' refers to standard reconstruction using observed parents and 'Cycled Back' refers to performing abduction on an observation's counterfactual and re-conditioning back on the observed parents. The x-axis shows the predicted label by the respective attribute classifier for each image.

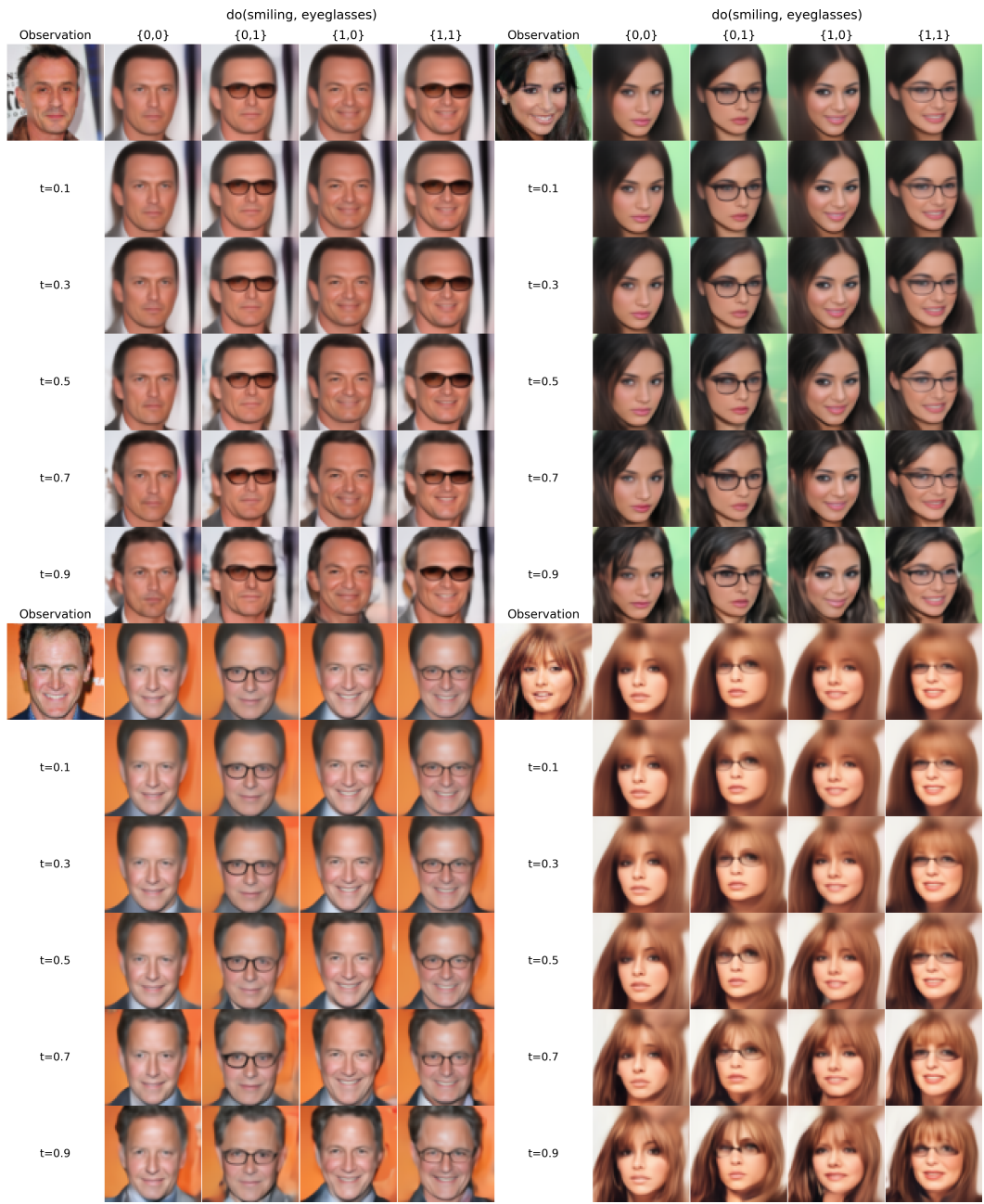

Figure 10: Traversing counterfactual latent space under different interventions and prior temperatures $t \in \{1e-5, 0.1, 0.3, 0.5, 0.7, 0.9\}$. We are visualising CelebA-HQ test set 'smiling' and 'eyeglasses' counterfactual combinations under partial abduction of the first 8 latent variables of our conditional VDVAE, where the remaining latents are sampled from the prior at different $t$'s per row.

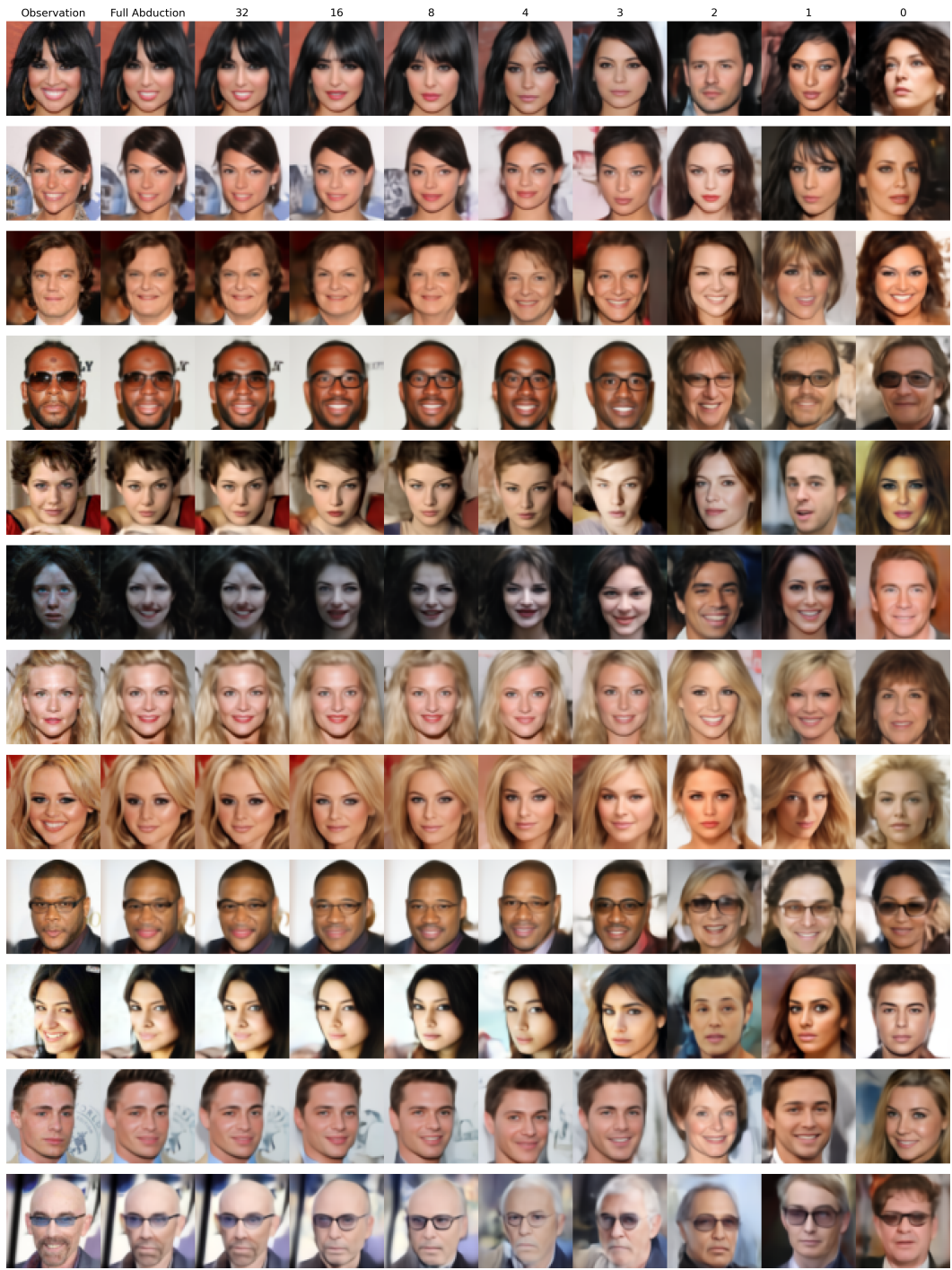

Figure 11: Visualising the evolution of 'smiling' CelebA-HQ counterfactuals from full abduction (42 latents) to partial abduction (32-1 latents) and finally random samples (0 latents fixed).

Figure 12: Visualising the evolution of 'eyeglasses' CelebA-HQ counterfactuals from full abduction (42 latents) to partial abduction (32-1 latents) and finally random samples (0 latents fixed).

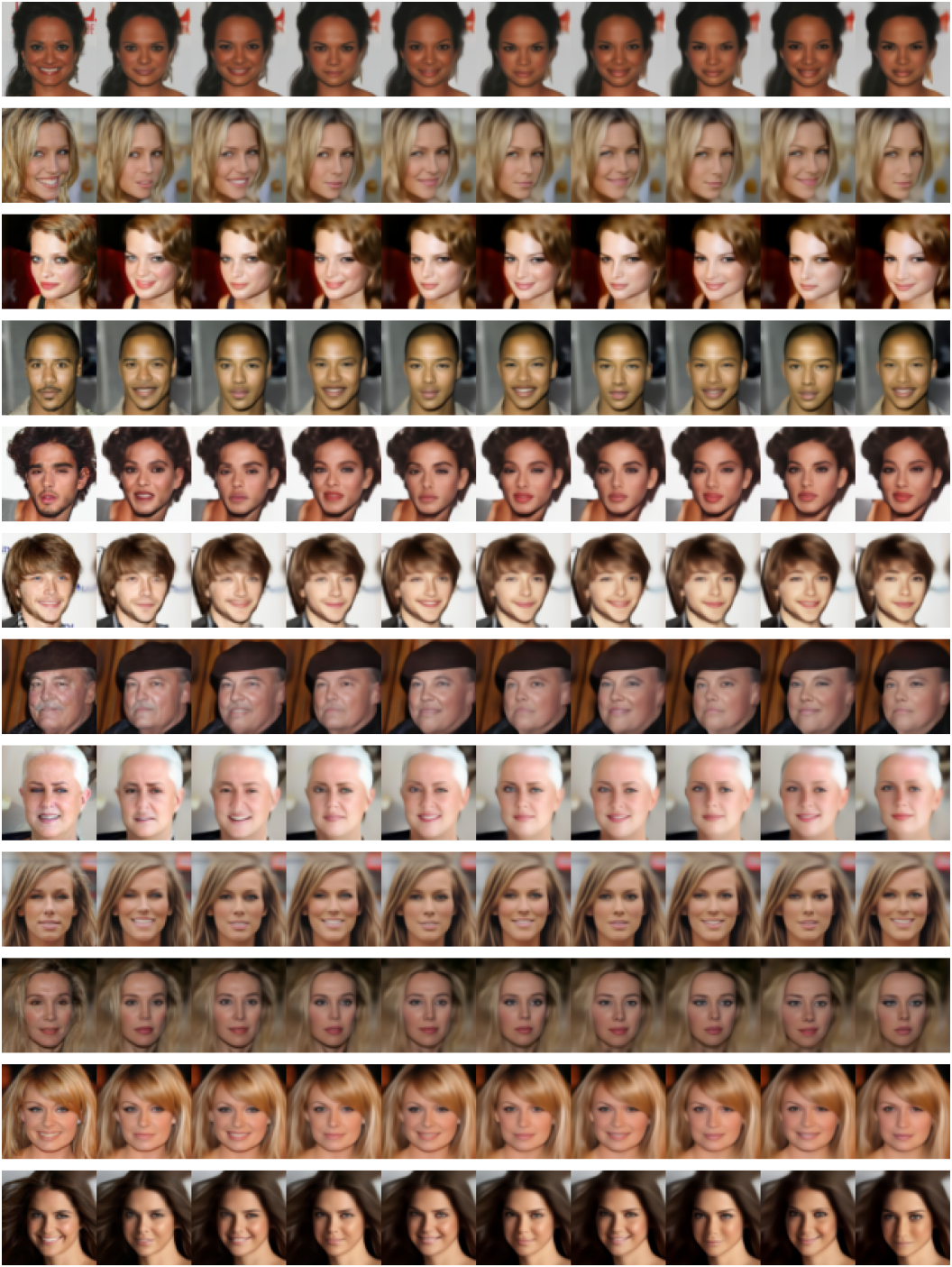

Figure 13: Visualising CelebA-HQ test set 'smiling' reversibility. First column from the left contains the observations, the second column contains 'smiling' counterfactuals under full abduction (42 latents), and the third column contains the first cycled back reversal. The process is repeated for the remaining columns, showing progressive degradation away from the original observation.